\documentclass[acmtog, authorversion]{acmart}
\usepackage{booktabs} % For formal tables
\usepackage{caption}
\usepackage{makecell}
\usepackage{multirow}
\usepackage{booktabs}
\usepackage{epstopdf}
\usepackage{marvosym}
\usepackage{graphicx}
\usepackage{amsmath}
\usepackage{colortbl}
\usepackage{changepage}
\usepackage{float}
\usepackage{stfloats}

\makeatletter
\let\@authorsaddresses\@empty
\makeatother

\renewcommand\footnotetextcopyrightpermission[1]{} % removes footnote with conference information in first column
\setcopyright{none}

\usepackage[ruled]{algorithm2e} % For algorithms

\SetAlFnt{\small}
\SetAlCapFnt{\small}
\SetAlCapNameFnt{\small}
\SetAlCapHSkip{0pt}

\acmJournal{TOG}
\begin{document}
% Title portion
\title{SPF-Portrait: Towards Pure Text-to-Portrait Customization with Semantic Pollution-Free Fine-Tuning}

% \author{Xiaole Xian}
% \authornote{Co-first authors. Listing order is random.}
% \affiliation{
%   \institution{Shenzhen University}
%   \city{Shenzhen}
%   \country{China}
% }
% \email{2310275030@email.szu.edu.cn}

% \author{Zhichao Liao}
% \affiliation{
%   \city{Shenzhen}
%   \country{China}
% }
% \email{liaozc23@mails.tsinghua.edu.cn}

\author{Xiaole Xian\textsuperscript{$\spadesuit$}\ }
\authornote{Co-first authors. Listing order is random.}
\author{Zhichao Liao\textsuperscript{$\heartsuit$}\ }
\authornotemark[1]
\affiliation{
  \institution{\textsuperscript{$\spadesuit$}Shenzhen University, \textsuperscript{$\heartsuit$}Tsinghua University}
  \city{\textsuperscript{$\spadesuit$}Shenzhen, \textsuperscript{$\heartsuit$}Beijing}
  \country{China}
}
% \email{2310275030@email.szu.edu.cn}
% \email{liaozc23@mails.tsinghua.edu.cn}

\author{Qingyu Li}
\author{Wenyu Qin}
\author{Pengfei Wan}
\affiliation{
  \institution{Kuaishou Technology}
  \city{Beijing}
  \country{China}
}
\email{liqingyu@kuaishou.com}
\email{qinwenyu@kuaishou.com}
\email{wanpengfei@kuaishou.com}

\author{Weicheng Xie}
\authornote{Joint corresponding authors.}
\author{LinLin Shen}
\affiliation{
  \institution{Shenzhen University}
  \city{Shenzhen}
  \country{China}
}
\email{wcxie@szu.edu.cn}
\email{llshen@szu.edu.cn}

\author{Long Zeng}
\authornotemark[2]
\author{Pingfa Feng}
\affiliation{
  \institution{Tsinghua University}
  \city{Shenzhen}
  \country{China}
}
\email{zenglong@sz.tsinghua.edu.cn}
\email{fengpf@tsinghua.edu.cn}

\begin{abstract}
Fine-tuning a pre-trained Text-to-Image (T2I) model on a tailored portrait dataset is the mainstream method for text-to-portrait customization.
However, existing methods often severely impact the original model’s behavior (e.g., changes in ID, layout, etc.) while customizing portrait attributes.
To address this issue, we propose \textbf{SPF-Portrait}, a pioneering work to purely understand customized target semantics and minimize disruption to the original model.
In our SPF-Portrait, we design a dual-path contrastive learning pipeline, which introduces the original model as a behavioral alignment reference for the conventional fine-tuning path.
During the contrastive learning, we propose a novel Semantic-Aware Fine Control Map that indicates the intensity of response regions of the target semantics, to spatially guide the alignment process between the contrastive paths.
It adaptively balances the behavioral alignment across different regions and the responsiveness of the target semantics.
Furthermore, we propose a novel response enhancement mechanism to reinforce the presentation of target semantics, while mitigating representation discrepancy inherent in direct cross-modal supervision.
Through the above strategies, we achieve incremental learning of customized target semantics for pure text-to-portrait customization.
Extensive experiments show that SPF-Portrait achieves state-of-the-art performance. 
Project page: \href{https://spf-portrait.github.io/SPF-Portrait/}{\textit{\textcolor{cyan}{https://spf-portrait.github.io/SPF-Portrait/}}}.
\end{abstract}


% \vspace{-2em}

\definecolor{base}{RGB}{67, 114, 194}
\definecolor{tar}{RGB}{176, 36, 24}
\begin{teaserfigure}
  \centering
  \includegraphics[width=0.99\textwidth]{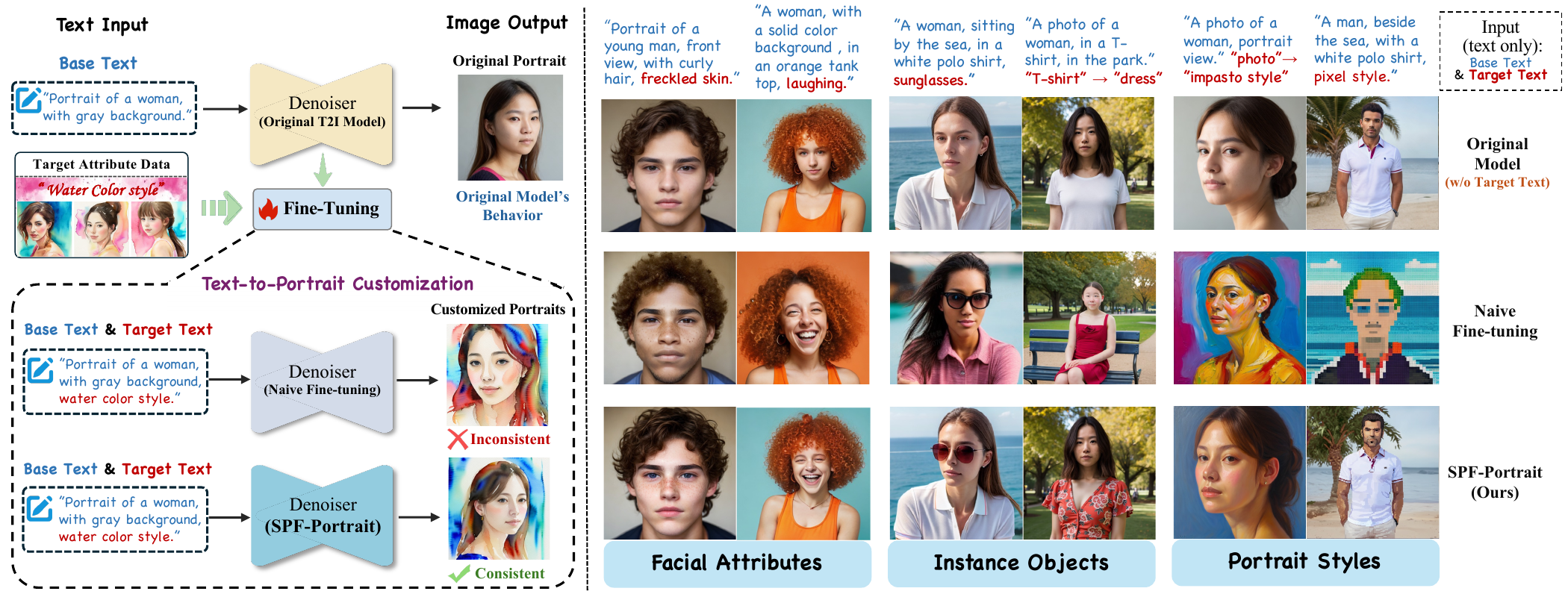}
  \vskip -1em
  \caption{\textbf{Left:} The paradigm of text-to-portrait customization. \textbf{Right:} Comparison of text-to-portrait customization across various dimensions. The original portrait, representing the original T2I model's behavior, is based solely on the \textbf{\textcolor{base}{Base Text}} as input.
  The customized portraits are generated by the fine-tuned model and are based on both the \textbf{\textcolor{base}{Base Text}} and \textbf{\textcolor{tar}{Target Text}} as inputs. 
  During the text-to-portrait customization, \textbf{our SPF-Portrait is able to achieve customized target semantics while maintaining consistency with the original model's behavior,  compared to naive fine-tuning.}}
  % \Description{1}
  \label{fig:teaser}
\end{teaserfigure}

\maketitle

% \input{samplebody-journals}

% \vspace{-1em}
\section{Introduction}
\label{sec:intro}

Fine-tuning pre-trained T2I diffusion models \cite{rombach2022high, ramesh2021zero, saharia2022photorealistic, esser2403scaling} offers an efficient approach for text-to-portrait customization \cite{huang2023collaborative, han2024face, he2024synfer}, which adapts the models to generate personalized target attributes.
However, as shown in Fig.~\ref{fig:teaser}, although conventional naive fine-tuning \cite{rombach2022high} can achieve target semantics, it has a significant impact on the original model’s behavior, such as altering the portrait's identity, posture, background, etc.
This is because, when the model learns the target semantics, the target semantics become entangled with redundant attributes \cite{hahm2024isometric} from the fine-tuning dataset. 
Consequently, while achieving the customized target semantics, the model not only generates the desired attributes but also inadvertently interferes with other original portrait attributes.
We refer to this phenomenon as \textbf{"Semantic Pollution"}, which is detrimental and often ignored.
This further indicates a non-incremental learning.
To address this issue, we propose SPF-Portrait, the first method to our knowledge that purely understands customized target semantics while eliminating semantic pollution in text-to-portrait customization. 
As shown in Fig.~\ref{fig:teaser}, our method is capable of stably performing well in customizing portrait attributes across various dimensions.

One line of previous research related to mitigating semantic pollution is PEFT-based methods \cite{hu2021lora, ding2023sparse, zhang2023adalora, borse2024foura, liu2023parameter}.
They minimize influence through low-rank adapter (e.g., LoRA and its variants \cite{ding2023sparse, zhang2023adalora, borse2024foura}) or orthogonal constraints \cite{qiu2023controlling, liu2023parameter}. 
However, their reliance on diffusion loss for implicit joint distribution modeling \cite{song2020score}, rather than understanding disentangled semantics, only allows for limited preservation of the original behavior.
Another line of work \cite{chefer2023attend, zhuang2024magnet, chen2024cat, cai2024decoupled, manas2024improving, liu2024improving, jiang2024frap} aims to purify the understanding of text embeddings and decouple attributes from each other. 
They enhance attribute independence through embedding-level decoupling (Magnet \cite{zhuang2024magnet}, TEBopt \cite{chen2024cat}) or attention regularization (Tokencompose \cite{wang2024tokencompose}).
While effective for instance-level generation (e.g., a cat or a dog), these methods fail when comes to refined attributes, such as hairstyles and skin textures.

Pure text-to-portrait customization manifests itself in generated portraits as introducing differences only by target attributes while maintaining consistency in unrelated attributes with the original model's outputs.
It requires achieving the following two objectives: 1) Effective adaptation of T2I models to target attributes, and 2) Faithful preservation of the original model's behavior. 
To this end, we propose the SPF-Portrait that incorporates an additional training stage after naive fine-tuning.
In this stage, we design a dual-path contrastive learning pipeline that introduces the frozen original model as the anchor of original behavior for the conventional fine-tuning path. 
During contrastive learning, we extract and constrain variant attention features and UNet features from the corresponding cross-attention layers in contrastive paths to align with the original performance.
We propose a novel Semantic-Aware Fine Control Map (SFCM) that accurately identifies the response regions of target semantics to spatially guide the alignment of these intermediate features.
This alignment process precisely aligns irrelevant attributes, avoiding suppression of target attribute and over-alignment.
Moreover, we propose a response enhancement mechanism for target semantics. 
By supervising the difference vectors of target semantics between the one-step prediction and the ground truth image, we mitigate the representational gaps inherent in direct cross-modal supervision and enhance the manifestation of target semantics.
Extensive experiments show that SPF-Portrait achieves state-of-the-art performance in preventing semantic pollution for pure text-to-portrait customization. In summary, our contributions are as follows:
\vspace{-0.25em}
% \vspace{-10pt}
\begin{itemize}
\item We propose SPF-Portrait, a dual-path contrastive learning pipeline, which is the pioneering work to address semantic pollution in text-to-portrait customization.
\item We introduce a novel Semantic-Aware Fine Control alignment process capable of preserving the original model's behavior while meticulously preventing over-alignment.
\item We design a response enhancement mechanism to improve the presentation of target semantics while alleviating representation gaps in direct cross-modal supervision.
\item Extensive quantitative and qualitative experimental results demonstrate the superiority of our SPF-Portrait.
\end{itemize}

%%%%%%%%%%%%%%%%%%%%%%%%%%%%%%%%%%%%%%%%%%%%%%%%%%%%%%%%%%%%%%%%%% Related Work

\vspace{-0.8em}

\section{Related Work}
\label{sec:related_work}

\noindent\textbf{Fine-tuning for T2I Diffusion Models.}
Numerous solutions \cite{zhang2023adding, wang2024instantid, ruiz2023dreambooth, huang2024realcustom, li2024anydressing, liao2024freehand} have improved existing T2I diffusion models \cite{rombach2022high, lin2024sdxl} in various aspects based primarily on fine-tuning.
Building upon the fine-tuning paradigm \cite{liao2025humanaesexpert,luo2024codeswap,luo2025object, wan2024grid}, PEFT-based methods \cite{wu2024difflora, zhang2023adalora, borse2024foura} rapidly adapt to new concepts by introducing additional parameters to the original model.
LoRA \cite{hu2021lora} achieves this through low-rank linear layers, while FouRA \cite{borse2024foura} based on LoRA further improves multi-concept integration by leveraging frequency domain learning.
Subsequent studies \cite{han2023svdiff, qiu2023controlling, liu2023parameter} further improve the preservation of prior knowledge during fine-tuning.
For instance, SVDiff \cite{han2023svdiff} fine-tunes only the singular values, the key parameters, via singular value decomposition. OFT \cite{qiu2023controlling} maintains the orthogonality of weight matrices, thereby preserving the hyperspherical energy of the pre-trained model.
Although they preserve pre-trained knowledge while adapting to new concepts, they overlook impure learning from relying solely on diffusion loss, causing new attributes to couple with irrelevant dataset attributes.

\noindent\textbf{Decoupling Generation of Diffusion Models.}
Efforts have also been made on decoupling control mechanisms, both between image-to-text conditions and within textual conditions, aiming to preventing the hinder to the textual control \cite{zhuang2024magnet, chen2024cat, qi2024deadiff, huang2024realcustom, xing2024csgo, gao2024styleshot, chang2024skews}.
To achieve the coupling within text, Magnet \cite{zhuang2024magnet} and TEBopt \cite{chen2024cat} analyze and optimize the condition embedding without additional training. However, while mitigating coupling at the instance level, they fail to correct the model’s deviation in understanding refined attributes.
RealCustom \cite{huang2024realcustom} dynamically adjusts image feature injection based on their impact on diffusion process, while DEADiff \cite{qi2024deadiff} tackles similar issues via a decoupling representation mechanism.
PuLID \cite{guo2024pulid} employs contrastive learning to prevent the injection of ID from disrupting the textual guidance to achieve decoupling.
However, these methods ignore the disruption from text conditions during fine-tuning with reference images.

\noindent\textbf{Distinction with Text-driven Image Editing Methods.}
\label{sec:editing}
The exceptional capability to adhere to base text enables our method to achieve end-to-end image manipulation \cite{brack2024ledits++, hoogeboom2023simple, instructcv} directly through T2I model, eliminating the need for additional editing pipelines.
While integrating text-driven editing methods \cite{deutch2024turboedit, wang2024belm, kim2022diffusionclip, ju2024brushnet, brooks2023instructpix2pix} into the T2I model pipeline can yield results comparable to ours.
For a image generated with T2I model, InstructPix2Pix \cite{brooks2023instructpix2pix} enables precise image manipulation through textual instructions by leveraging a conditioned diffusion model trained on paired image editing datasets. Similarly, DiffusionCLIP \cite{kim2022diffusionclip} and Asyrp \cite{kwon2022diffusion}, inspired by GAN-based methods \cite{alaluf2022hyperstyle}, utilize a local directional CLIP loss \cite{baykal2023clip} between images and text to manipulate specific attributes.
However, the task of our work lies in preventing new textual attributes from disrupting T2I models, which fundamentally differs from the goal of I2I editing models that focus on image manipulation.

\section{Methodology}

Our SPF-Portrait improves naive fine-tuning by introducing an additional training stage.
\textbf{In the first stage}, we employ naive fine-tuning to strive for the preliminary response to target semantics without considering the contamination to the original model.
\textbf{In the second stage}, we design a Dual-path Contrastive Learning approach (Sec.~\ref{sec:Pipeline}) that introduces the frozen original model along with the fine-tuning path.
During contrastive learning, we propose the Semantic-Aware Fine-Control Map to guide alignment with the original model's behavior (Sec.~\ref{sec:Alignment}) and design the Response Enhancement mechanism for target semantics (Sec.~\ref{sec:Response}).

\subsection{Preliminary}
\label{sec:Preliminary}

\noindent\textbf{Diffusion Models.} T2I diffusion models generate images based on text input through a forward diffusion process and a reverse denoising process \cite{ho2020denoising, saharia2022photorealistic}. 
The diffusion process follows the Markov chain to transform an image sample
$x_{0}$ into noisy samples $x_{1:T}$ by adding Gaussian noise $\epsilon$ over $T$ steps.
The denoising process employs a denoising model $\epsilon_\theta$ to predict the added noise using $x_{t}$, $t$, and textual conditions $y$ as inputs, where $\theta$ denotes the learnable parameters and $t\in[0,T]$ is the diffusion process timestep. The optimization process can be described as:
\begin{align}
\label{eq:diff_loss}
    \mathcal{L}_{diff}=\mathbb{E}_{x_0,\epsilon\thicksim\mathcal{N}(0,1),t}(\|\epsilon-\epsilon_\theta(x_t,t, E)\|_2^2),
\end{align}
where $E=\tau_{text}(y)$ is textual features, obtained from the textual conditions $y$ encoded by the text encoder $\tau_{text}$.

\subsection{Dual-Path Contrastive Learning Pipeline}
\label{sec:Pipeline}

\begin{figure}[!t]
  \centering
  \includegraphics[width=0.9\linewidth]{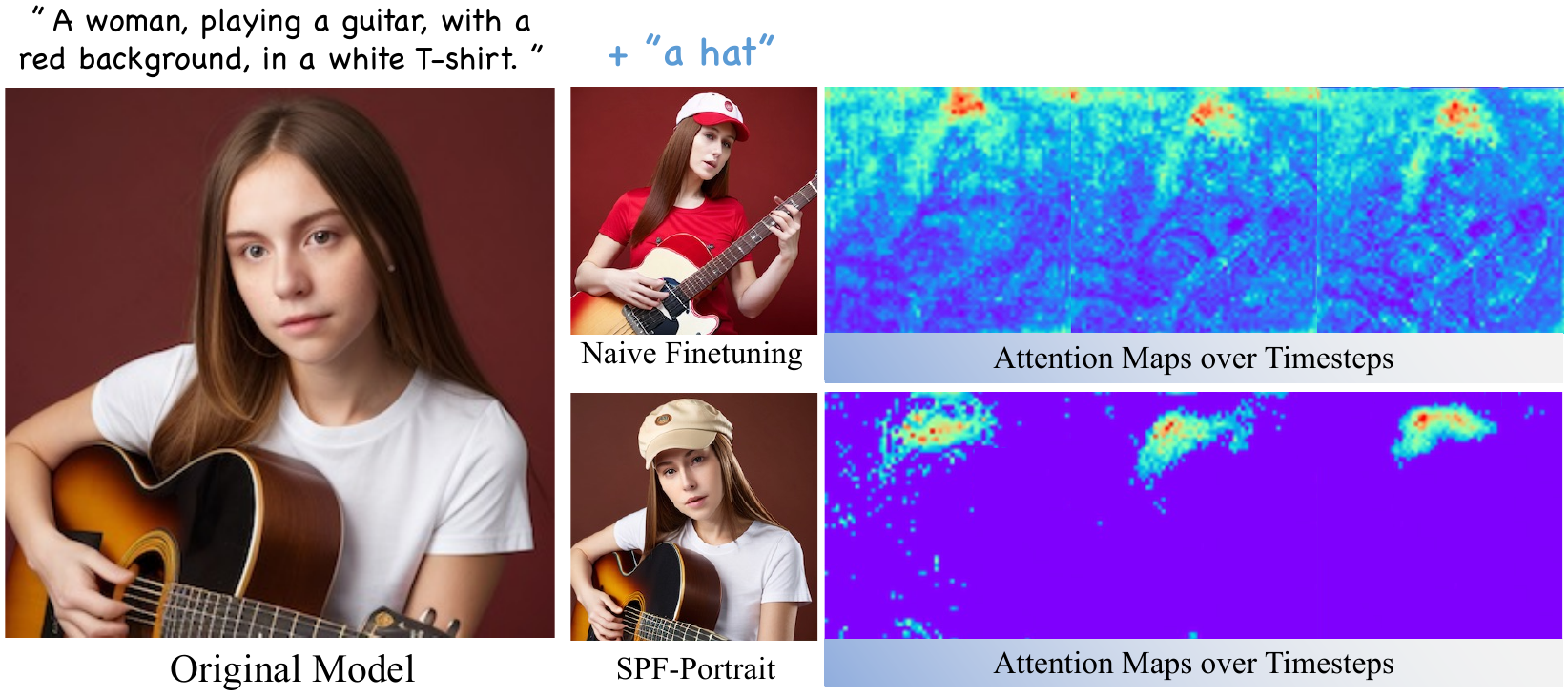}
  \vskip -1em
  \caption{
  \textbf{Visualization of the Attention Map.} The salient regions directly reflect response intensity to the target semantics "a hat". % Brighter regions indicate higher attention.
  }
  \label{fig:attn_map}
  \vskip -1em
\end{figure}

Although the first stage of training, a naive fine-tuning, can initially achieve the adaptation of T2I models to target attributes.
However, as shown in Fig.~\ref{fig:teaser}, it will severely affect the behavior of the original model.
We visualize the attention map \cite{vaswani2017attention} of target text after naive fine-tuning in Fig.~\ref{fig:attn_map} to diagnose this limitation. 
The response regions of the target semantics are extended to unrelated areas, interfering with other attributes, which is caused by semantic pollution during fine-tuning.
To address this issue, we design an additional training stage that utilizes a dual-path contrastive learning pipeline.
Specifically, the proposed dual paths including:
(i) \textbf{Reference Path} comprises a frozen model initialized from the original pre-trained T2I model. In contrastive learning, it only takes $E^{ref}_{base} = \tau_{text}(y_{base})$ as input, serving as a stable reference on behalf of the original model's behavior; 
and (ii) \textbf{Response Path} includes a model initially resumed from the first stage. During contrastive learning stage, it takes complete text (i.e., $y_{base}$ and $y_{tar}$) as input:
\begin{equation}
\begin{split}
\label{eq:text_feat}
    E_{base}^{res} &= \tau_{text}([y_{base}, y_{tar}])|_{y_{base}},\\ E_{tar} &= \tau_{text}([y_{base}, y_{tar}])|_{y_{tar}},
\end{split}
\end{equation}
where $[y_{base}, y_{tar}]$ represents the concatenated text prompt. $E_{base}^{res}$ and $E_{tar}$ represent the encoded feature segments corresponding to $y_{base}$ and $y_{tar}$ portions respectively.
By contrastive learning between dual paths, we specifically design a Semantic-Aware Fine Control alignment process to maintain the original model's behavior and an response enhancement mechanism for target semantics.

\definecolor{base}{RGB}{67, 114, 194}
\definecolor{tar}{RGB}{176, 36, 24}
\begin{figure*}[!ht]
  \centering
  \includegraphics[width=1.0\linewidth]{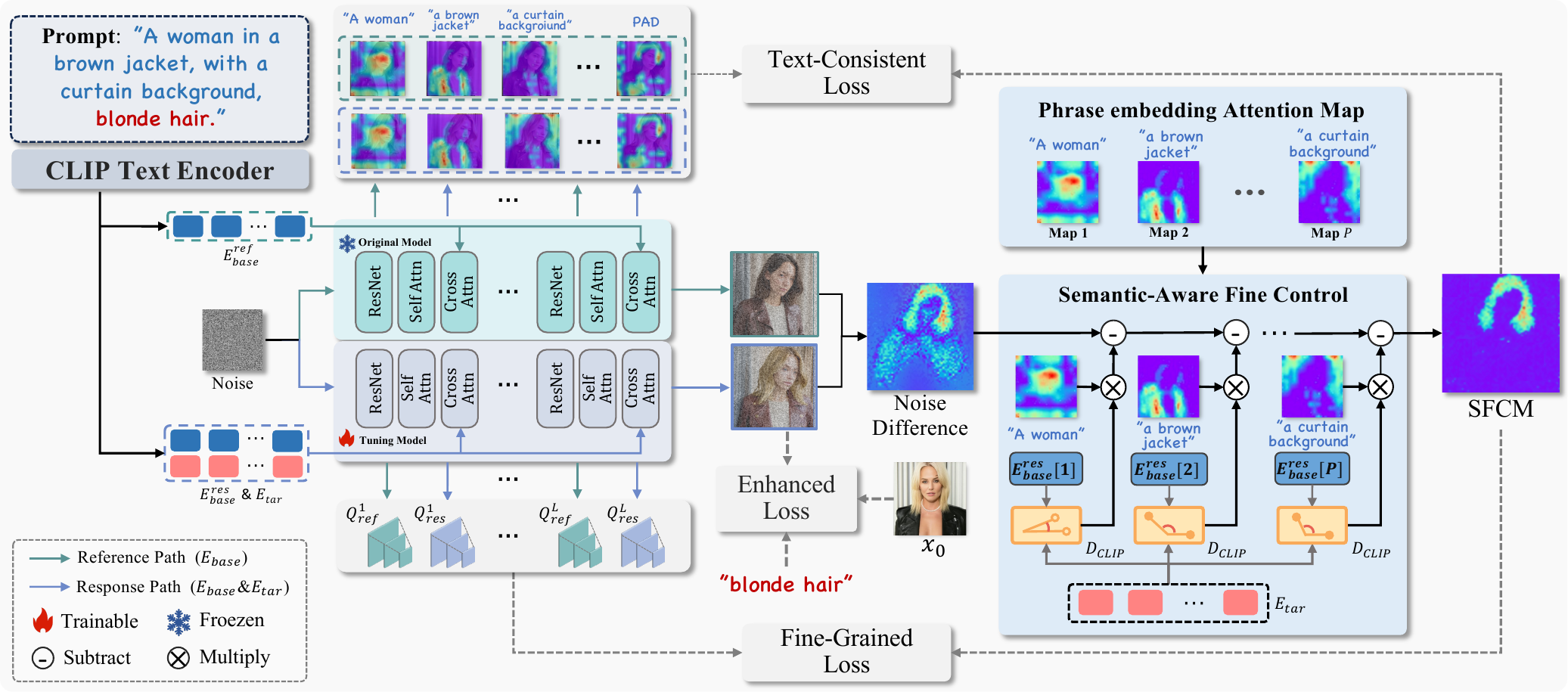}
  \vskip -1em
  \caption{\textbf{The Dual-Path Contrastive Learning Pipeline of SPF-Portrait.} 
  The text in \textbf{\textcolor{base}{blue}} is the \textbf{\textcolor{base}{Base text}}, while those in \textbf{\textcolor{tar}{red}} is the \textbf{\textcolor{tar}{Target text}}. Reference Path takes only \textbf{\textcolor{base}{Base text}}
  as input, while Response Path takes complete text (\textbf{\textcolor{base}{Base text}} \& \textbf{\textcolor{tar}{Target text}}) as input.
  }
  \label{fig:pipeline}
\end{figure*}

\subsection{Semantic-Aware Fine Control Alignment}
\label{sec:Alignment}

In this section, we provide a detailed presentation of our novel Semantic-Aware Fine Control alignment process. This process first extracts the attention features $\mathcal{F}_{ref}$ and $\mathcal{F}_{res}$ from the reference path and response path.
These features are derived from a variant of the standard attention mechanism, i.e., Attention $(K, Q, Q)$.
They represent the response of the UNet features $Q_{ref}$ and $Q_{res}$ to the base textual features $E_{base}$, where $Q_{ref}$ and $Q_{res}$ are features from the corresponding UNet’s cross-attention layer in the contrastive paths.
By constraining the similarity between the attention features $\mathcal{F}_{ref}$ and $\mathcal{F}_{res}$ from each cross-attention layer, this process encourages the representation of the base text in the response path to approach the behavior of the original model as:
\begin{align}
    \left.\left.\left.\left\{
    \begin{array}
        {l}\mathcal{F}_{ref} = \mathrm{Softmax}(\frac{K_{ref}(E^{ref}_{base})\ \ Q_{ref}^T}{\sqrt{d}})Q_{ref}, \\
        \mathcal{F}_{res} = \mathrm{Softmax}(\frac{K_{res}(E^{res}_{base})\ \ Q_{res}^T}{\sqrt{d}})Q_{res},\\
        \mathcal{L}_{\text{text-consistent}}=\sum_{j=1}^{L}\left\| \mathcal{F}^{j}_{ref} - \mathcal{F}^{j}_{res} \right\|_2,\
    \end{array}\right.\right.\right.\right.
    \label{eq:consistent}
\end{align}
where $K_{ref}$ and $K_{res}$ denotes the key of $E^{ref}_{base}$ and $E^{res}_{base}$ in dual-path. $L$ represents the attention layer number of the denoising model.

To enhance consistency in fine-grained content, we further constrain the UNet features $Q$ from contrastive paths, which contains comprehensive information on local details and global structure \cite{chung2024style, mo2024freecontrol}. This is formulated as:
\begin{align}
    \mathcal{L}_{\text{fine-grained}}=\sum_{j=1}^{L}\left\|Q^{j}_{ref}-Q^{j}_{res}\right\|_2.
    \label{eq:fine-graned}
\end{align}

\begin{figure}[!t]
  \centering
  \includegraphics[width=1.0\linewidth]{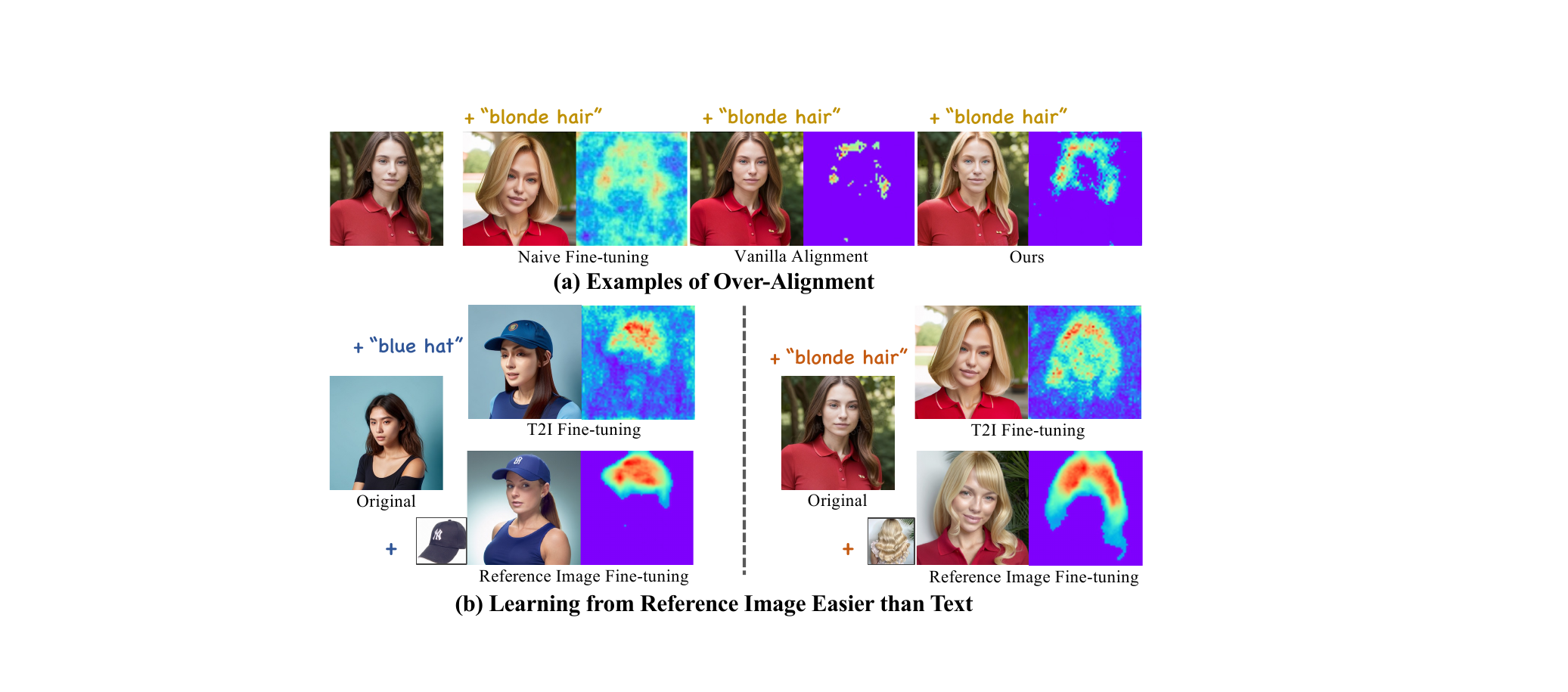}
  \vskip -1em
  \caption{
  \textbf{Analysis of Alignment Process. }\textbf{(a)} Vanilla alignment results in the over-alignment with original portrait.
  \textbf{(b)} For the same customization attribute, reference image-based fine-tuning offers a more distinct target response region than T2I fine-tuning.
  % Learning textual concepts through fine-tuning coupled with more entangled irrelevant feature compared to learning from reference image representations. 
  }
  \vskip -1em
  \label{fig:attn_map_abla}
\end{figure}

Although such a contrastive alignment effectively prevents the impact of the original model (e.g., in reference image-based customization tasks \cite{guo2024pulid}), this vanilla alignment of intermediate features in text-driven generation suppresses the response intensity of target semantics, as shown in Fig.~\ref{fig:attn_map_abla} (a). This causes the customized portrait to overly align with the original portrait.
As shown in Fig.~\ref{fig:attn_map_abla} (b), the fundamental distinction lies in the learning objectives.
Since the reference image is inherently decoupled from text and represents a more concrete condition, it allows the model to have well-defined objectives to consult, thereby having negligible disturbance on target attribute performance.
In contrast, the semantic boundaries between textual concepts are ambiguous, which finally amplifies the influence.
To address this more challenging issue, we propose a Semantic-Aware Fine Control Map (SFCM) that spatially guides the alignment process to be implemented in the appropriate regions, minimizing its disturbance on the target response.
Specifically, during alignment training, the spatial difference in noise predictions between contrastive paths can serve as prior knowledge for target response, forming a soft map $\mathcal{M}$ as:
\begin{equation}
\mathcal{M} = |\epsilon_\theta(x_t,t,E^{ref}_{base}) - \epsilon_{\theta^{'}}(x_t,t,[E^{res}_{base}, E_{tar}])|,
\end{equation}
where the $\epsilon_{\theta^{'}}$ and $\epsilon_\theta$ represent the prediction in both response and reference paths, respectively, while $\theta^{'}$ denoting the  learnable parameters.
As previously analyzed, Semantic Pollution causes the target response regions to diffuse into areas of other attributes, making the noise difference $\mathcal{M}$ unable to precisely characterize the target response regions.
Inspired by the insight that if a phrase in base text exhibits low semantic relevance to target text, the regions highlighted by this phrase should be excluded from the $\mathcal{M}$, we design the Semantic-Aware process to refine the soft map. 
For the input base text in response path, we split it into multiple phrases, as shown in "Phrase embedding Attention Map" of Fig. \ref{fig:pipeline}.
Concretely, for each phrase feature $E^{res}_{base}[i],~i = \{1,2, \cdots, P\}$and $P$ is the total number of phrase in base text, we compute its mean of the cross-attention maps across all the UNet layers to localize highlighted regions $\overline{A}_{base}[i]$ as:
\begin{align}
        \overline{A}_{base}[i] = \frac{1}{L} \sum_{j=1}^{L}{(A_{base}^{j}[i] )},
\label{feature} 
\end{align}
where $A_{base}^{j}[i]$ represents the attention map of the $i$-th phrase embedding $E^{res}_{base}[i]$ from the $j$-th layer. Subsequently, to quantify the relevance of exclusion, we leverage the representation capabilities of CLIP to calculate the similarity between $E_{tar}$ and each $E^{res}_{base}[i]$. 
We then weight the $\overline{A}_{base}[i]$ based on the similarity, which used to refine the soft map $\mathcal{M}$, as expressed below:
\begin{equation}
\label{feature}
\begin{split}
    \widehat{\mathcal{M}} = \mathcal{M} - &\sum_{i=1}^{P}\overline{A}_{base}[i] \cdot (1-\gamma (i) ), \\
    \gamma (i) =& D_{CLIP} (E^{res}_{base}[i], E_{tar} ),
\end{split}
\end{equation}
where $D_{CLIP}$ represent the cosine similarity in CLIP embedding space. All attention maps are upsampled at a resolution of 64 $\times$ 64 as the same as noise map. $\widehat{\mathcal{M}}$ is our final SFCM, as shown in Fig.~\ref{fig:pipeline} and Fig.~\ref{fig:attn_map_abla} (a), it represents the precise target response regions and effectively prevents over-alignment by guiding the alignment process.

Therefore, the alignment constraints in Eq. \ref{eq:consistent} and Eq. \ref{eq:fine-graned} can be modified as follow:
\begin{equation}
\begin{array}{cc}
      \mathcal{L}_{M\mathrm{-tex}} = \sum_{j=1}^{L}\left\| (\mathcal{F}^{j}_{ori} - \mathcal{F}^{j}_{ft} ) \odot (1 - \widehat{\mathcal{M}}) \right\|_2, \\
      % \addlinespace[2pt]
      \mathcal{L}_{M\mathrm{-fine}}=\sum_{j=1}^{L}\left\| (Q^{j}_{ori} - Q^{j}_{ft}) \odot (1 - \widehat{\mathcal{M}}) \right\|_2, \\
\end{array}
\end{equation}
where $\odot$ denotes the hadamard product.

\subsection{Response Enhancement via Difference Vectors}
\label{sec:Response}
In text-to-portrait customization, an excellent response to the target semantics is essential for success.
Therefore, to reinforce the model's comprehension of the target attribute, we devise a response enhancement mechanism to improve the presentation of the target semantics.
Specifically, we introduce a difference vector $\Delta$, represented by the difference between the vectors of the CLIP textual space and the CLIP visual space \cite{abdelfattah2023cdul, xue2022clip}.
By introducing the ground truth image ${x}_{0}$ (a image with target attribute), we separately calculate the difference vector $\Delta({x}_{0}, E_{tar})$ between the target text and ground truth image ${x}_{0}$, as well as the difference vector $\Delta(\hat{x}_{0}, E_{tar})$ between the target text and the one-step prediction $\hat{x}_{0}$, formulated as:
\begin{equation}
\begin{split}{}
     & \Delta(\hat{x}_{0}, E_{tar}) = \tau_{vision}(\hat{x}_{0})-\tau_{text}(E_{tar}),\\
     & \Delta(x_{0}, E_{tar}) = \tau_{vision}(x_{0})-\tau_{text}(E_{tar}),\\
     \hat{x}_{0} =& \frac{\widehat{x}_{t}}{\sqrt{\bar{\alpha}_t}}-\frac{\sqrt{1-\bar{\alpha}_t}\epsilon_\theta(\widehat{x}_{t},t, \tau_{text}([E^{res}_{base}, E_{tar}] )}{\sqrt{\bar{\alpha}_t}},\\
\end{split}
\label{vec}
\end{equation}
where the $\tau_{vision}$ and $\tau_{text}$ denote the CLIP vision and text encoder, respectively, while $\hat{x}_{0}$ denotes the one-step prediction of $x_{t}$ in $t$-th timestep. Then, we constrain their similarity to enhance the response of the target semantics as:
\begin{equation}
\begin{split}
\!\mathcal{L}_{enhanced} \!= \!1 \!-\! D_{CLIP}(\Delta(\hat{x}_{0}, \!E_{tar}), \Delta(x_{0}, \!E_{tar}) ).
\end{split}
\label{clip_loss}
\end{equation}

\begin{figure}[!t]
  \centering
  \includegraphics[width=1.0\linewidth]{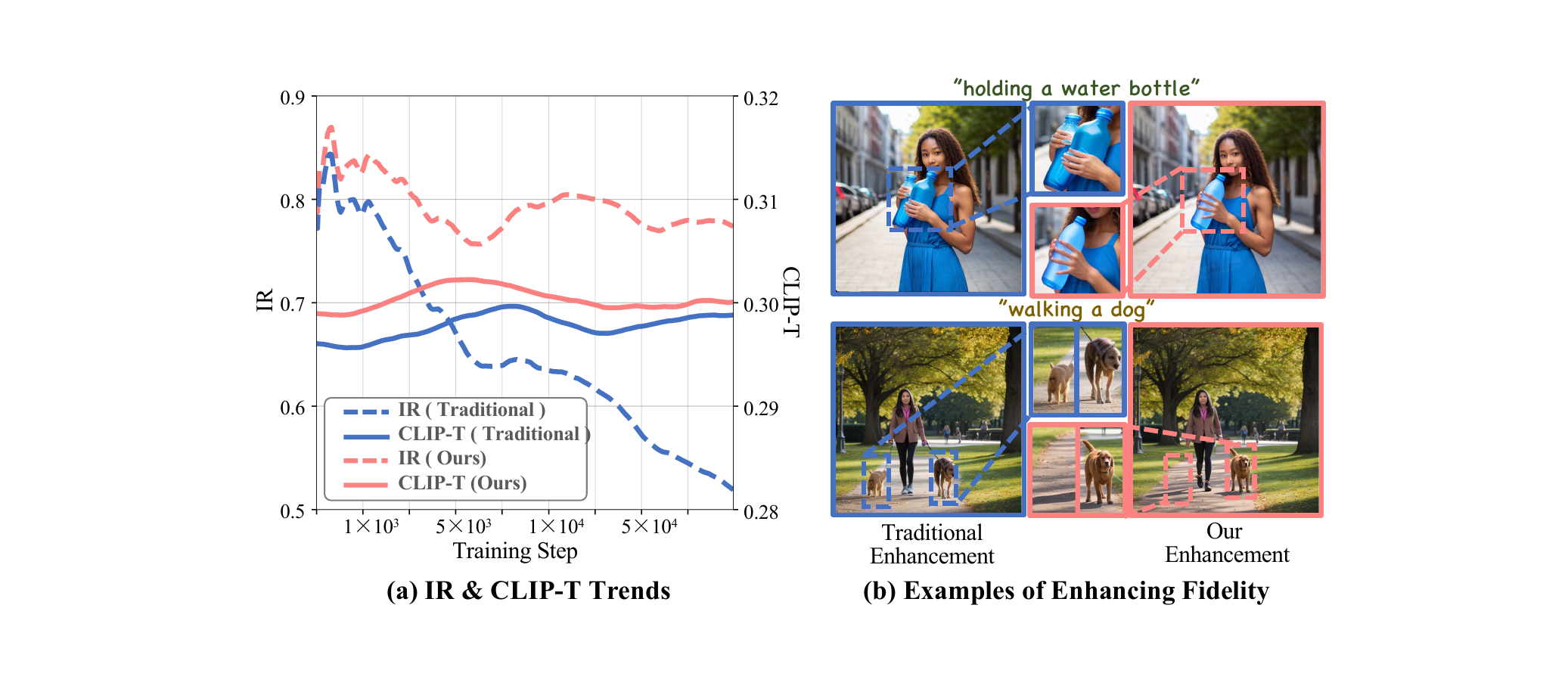}
  \vskip -1em
  \caption{
  \textbf{Comparison with Traditional Supervision on Image Fidelity.}
  \textbf{(a)} illustrates the trend of Image-Reward (IR) and CLIP Score (CLIP-T) across training steps. Image-Reward \cite{xu2023imagereward} is a metric used to evaluate image fidelity.
  \textbf{(b)} displays samples from traditional method \cite{avrahami2022blended} and ours.
  }
  \label{fig:enhance}
\end{figure}

\begin{figure*}[tb]
  \centering
  \includegraphics[width=0.95\linewidth]{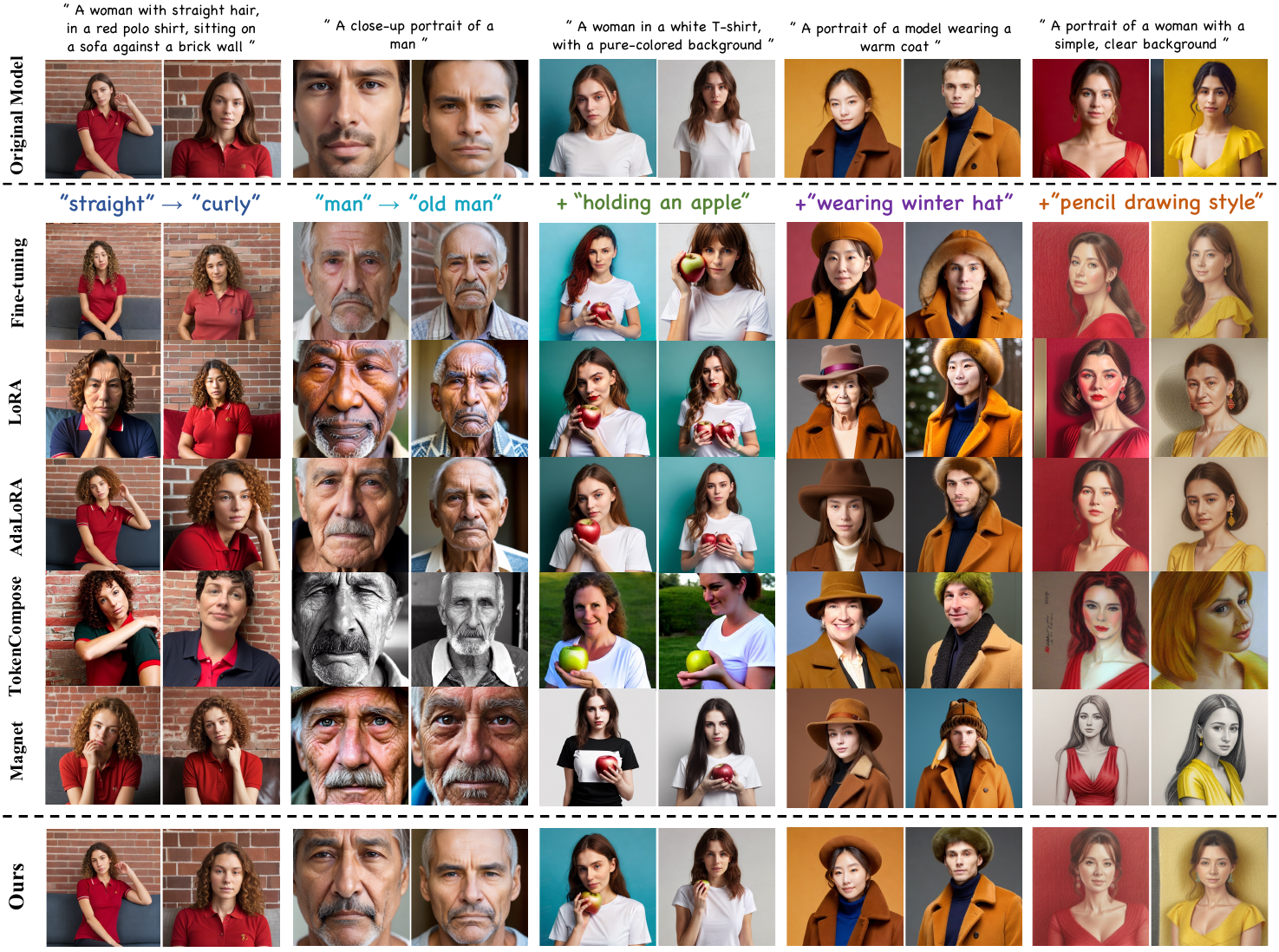}
  \vskip -1em
  \caption{\textbf{Qualitative Comparisons with SOTA methods.} We compare ours with naive fine-tuning \cite{rombach2022high}, PEFT-based methods (LoRA \cite{hu2021lora}, AdaLoRA \cite{zhang2023adalora} ) and the decoupled methods (Tokencompose \cite{wang2024tokencompose}, Magenet \cite{zhuang2024magnet}). 
  % Please zoom in for more details.
  % Our approach not only achieves the target semantics, but also better preserves the information from the original model. 
  }
  \label{fig:sota}
\end{figure*}

Unlike previous work \cite{avrahami2022blended,kim2022diffusionclip} that directly applies cross-modal supervision in CLIP space by employing the target text to supervise the one-step prediction \cite{yin2024one}, formulated as:
\begin{equation}
    \begin{split}
        \mathcal{L}_{clip} =& 1- D_{CLIP}(\tau_{vision}(\hat{x}_{0})-\tau_{text}(E_{tar})).\\
        % \hat{x}_{0} = \frac{\widehat{x}_{t}}{\sqrt{\bar{\alpha}_t}}-&\frac{\sqrt{1-\bar{\alpha}_t}\epsilon_\theta(\widehat{x}_{t},t, \tau(E_{base}, E_{tar})  )}{\sqrt{\bar{\alpha}_t}},
    \end{split}
\label{eq:ori_clip_loss}
\end{equation}
our approach reformulates the optimization objective into difference vectors rather than image-text similarity in Eq. \ref{eq:ori_clip_loss}.
Directly cross-modal supervision overlooks the modality representation gap, causing the model to overfit the textual description during optimization and neglecting the visual fidelity of the result image.
As illustrated in Fig.~\ref{fig:enhance}, it ultimately leads to degradation in the quality of the generated images.
In contrast, we provide an effect similar to supervision within the same modality by using the difference between cross-modal vectors, mitigating the representation discrepancy inherent in direct cross-modal supervision. It simultaneously enhances the response to target semantics while improving the fidelity and coherence of the image.

Finally, the overall optimization objective can be represented as:
\begin{equation}
\mathcal{L}_{SPF}= \mathcal{L}_{diff} + \underbrace{\lambda_{1}\mathcal{L}_{\mathrm{M\mathrm{-text}}} +\lambda_{2}\mathcal{L}_{M\mathrm{-fine}}}_{\mathrm{alignment}} + \underbrace{\lambda_{3}\mathcal{L}_{enhacned}}_{\mathrm{\: response}} ,
% \raisetag{30pt}
\label{eq:final_loss}
\end{equation}
where $\lambda_{1}$, $\lambda_{2}$ and $\lambda_{3}$ are the hyperparameters.

\section{Experiments}
\label{sec:Experiment}

\subsection{Experimental Setup}

\noindent{\textbf{Implementation Details. }}We adopt the Stable Diffusion v1.5 model \cite{rombach2022high} with 
Realistic\_Vision\_V4.0 checkpoints.
The hyperparameters $\lambda_{1}$, $\lambda_{2}$ and $\lambda_{3}$ are set to 0.2, 0.1 and 0.6.
More details about experiments are provided in the \textit{Appendix}.

\noindent{\textbf{Dataset. }}Our training set contains 230K diverse portraits with new attributes (e.g., skin textures, hairstyles), captioned by GPT-4o \cite{achiam2023gpt} and Cambrian-1 \cite{tong2024cambrian}.
For evaluation, we create a test set of 5K triples, each with: (1) an original caption, (2) its corresponding original portrait generated using Realistic\_Vision\_V4.0, and (3) a target caption of customized attributes.

\noindent{\textbf{Evaluation Metrics. }}We evaluate three key aspects: (1) preservation of the original model's behavior, (2) responsiveness to target semantics, and (3) overall image quality. Concretely, we employ FID \cite{heusel2017gans}, LPIPS \cite{zhang2018unreasonable}, identity similarity (ID), CLIP Image Score (CLIP-I) \cite{radford2021learning}, and segmentation consistency \cite{kirillov2023segment} (Seg-Cons) to measure the consistency between the original and customized portraits. We use the CLIP Score (CLIP-T) \cite{radford2021learning} to evaluate responsiveness to target semantics. For overall image quality assessment, we use HPSv2 \cite{wu2023human} and MPS \cite{Zhang_2024_CVPR}.

\begin{figure*}[!t]
  \centering
  \includegraphics[width=0.9\linewidth]{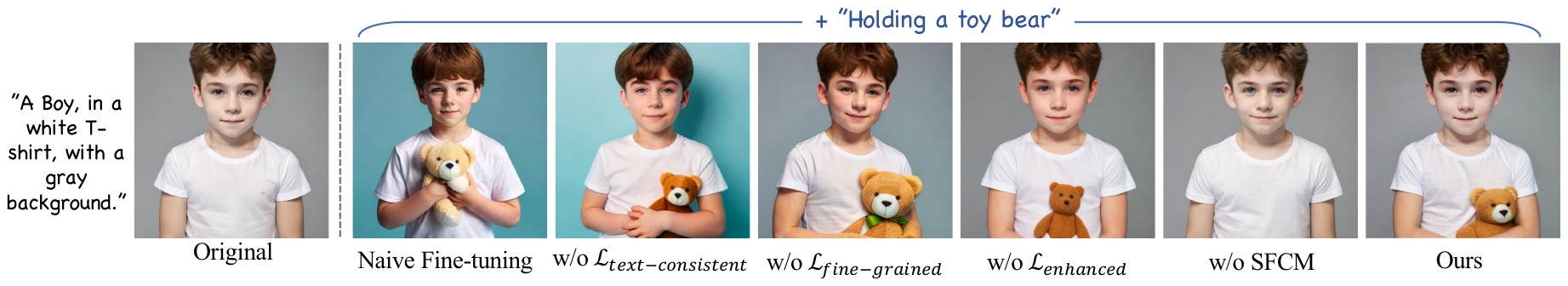}
  \vskip -1em
  \caption{\textbf{Qualitative Ablation Study.} We independently ablate the proposed loss and the SFCM mechanism.}
  \label{fig:ablation}
  \vskip -1em
\end{figure*}

\subsection{Qualitative Evaluation}
\textbf{Comparison with SOTAs.} We qualitatively comparison of our method with the SOTA approaches, including PEFT-based methods such as LoRA \cite{hu2021lora} and AdaLoRA \cite{zhang2023adalora},  decoupled text embedding methods like TokenCompose \cite{wang2024tokencompose} and Magnet \cite{zhuang2024magnet}, as well as naive fine-tuning.
We compare with them on diverse customized attributes, such as age, image style, and clothing. For each target attribute, we evaluate two cases under different random seeds.
As shown in Fig.~\ref{fig:sota} and Fig.~\ref{fig:appendix_sota}, although LoRA \cite{hu2021lora} and AdaLoRA \cite{zhang2023adalora} tend to retain original behavior in some cases, their performance is extremely unstable and poor in detail alignment.
For instance, in row 3, column 3, there is a noticeable change in identity, whereas in row 4, column 2, the pose of portrait has transformed completely.
Magnet \cite{zhuang2024magnet} and TokenCompose \cite{wang2024tokencompose} naively follow the input text conditions entirely, ignoring the preservation of the original model's behavior across all test cases.
For example, in row 6 \& 7, column 9, the customization of ``pencil drawing style" results in a total alteration of the portrait.
In contrast, our method purely customizes target attributes while preserving the original model's behavior in aspects such as background, pose, and identity. It demonstrates our approach effectively address semantic pollution during fine-tuning. 

\noindent\textbf{More Extensions.} We provide two more extensions of our SPF-Portrait: 1) As shown in Fig~\ref{fig:appendix_con}, our method reliably performs excellently in continuous replacements and additions of target text in text-to-portrait customization. 2) In Fig~\ref{fig:appendix_gs}, we demonstrate the strong potential of extending our method to the General T2I domain.

\definecolor{tb-blue}{RGB}{196, 228, 252}
\begin{table*}[ht]
\centering
\caption{
\textbf{Quantitative Comparison Results. } 
% 有灰色背景的行表示我们的消融实验，没有的表示需要对比的SOTA方法。
Rows without color represent comparisons with SOTA methods, while blue rows indicate our ablation experiments.
In our specific pairwise comparison, unlike general image generation, lower FID values reflect greater consistency with the original model's behavior.
Notably, the underlined values in "Ours (w/o SFCM)" are unusually low because the generated portraits overly align with the original portraits.
% It is notable that ``Ours (w/o SFCM)"中有下划线的数字表现出异常的低，这是因为它生成的肖像与原肖像过度对齐。
}
\label{tab:sota_comparsion}
\resizebox{1.0\textwidth}{!}
{
\begin{tabular}{lcccccccc}
\toprule
\multicolumn{1}{c}{\multirow{2}{*}{\textbf{Method}}} & \multicolumn{5}{c}{\textbf{Preservation}} & {\textbf{Responsiveness}} & \multicolumn{2}{c}{\textbf{Overall}} \\
\cmidrule(r){2-6} \cmidrule(r){7-7} \cmidrule(r){8-9} 
& FID ($\downarrow$) & LPIPS ($\downarrow$) & ID ($\uparrow$) & CLIP-I ($\uparrow$) & Seg-Cons ($\uparrow$) &  CLIP-T ( $\uparrow$) & HPSv2 ($\uparrow$)   & MPS($\uparrow$)  \\ 

\midrule

\multicolumn{1}{l|}{Naive Fine-Tuning \cite{rombach2022high}}  & 20.41              & 0.57                 & 0.21            & 0.63                & 57.77                 & 0.24                 & 0.21                  & 0.67      \\
\multicolumn{1}{l|}{AdaLoRA \cite{zhang2023adalora}}  & 7.38               & 0.40                 & 0.39            & 0.80                & 64.86                 & 0.23                 & 0.24                  & 1.10   \\
\multicolumn{1}{l|}{LoRA \cite{hu2021lora}}  & 9.82               & 0.38                 & 0.52            & 0.71                & 58.37                 & 0.27                 & 0.23                  & 1.21   \\ 

\midrule

\multicolumn{1}{l|}{TokenCompose \cite{wang2024tokencompose}} & 10.93              & 0.41                 & 0.32            & 0.81                & 40.22                 & 0.27                 & 0.24                  & 0.71     \\
\multicolumn{1}{l|}{Magnet \cite{zhuang2024magnet}}  & 18.92              & 0.48                 & 0.38            & 0.61                & 32.87                 & 0.26                 & 0.26                  & 0.97     \\

\midrule 
\rowcolor{tb-blue!40}
\multicolumn{1}{l|}{\textbf{Ours}} & \textbf{4.50} & \textbf{0.35} & \textbf{0.55} & \textbf{0.83} & \textbf{75.74} & \textbf{0.30} & \textbf{0.28} & \textbf{1.49} \\

\bottomrule

\rowcolor{tb-blue!25}\multicolumn{1}{l|}{Ours (w/o $\mathcal{L}_{\mathrm{text- consistent}}$) }  & 4.97  & 0.39 & 0.48 & 0.60 & 61.39 & 0.28 &  0.23 &  1.13  \\
\rowcolor{tb-blue!20}\multicolumn{1}{l|}{Ours (w/o $\mathcal{L}_{fine-grained}$)  }& 6.74  & 0.42 & 0.32 & 0.71 & 41.62 & 0.27  &   0.21 &   1.22  \\
\rowcolor{tb-blue!12}\multicolumn{1}{l|}{Ours (w/o $\mathcal{L}_{enhanced}$) }        & 4.52  & 0.37 & 0.49 & 0.81 & 74.38 & 0.22   & 0.23  & 1.40    \\
\rowcolor{tb-blue!5}\multicolumn{1}{l|}{Ours (w/o SFCM) }                            & \underline{4.13}  & \underline{0.14} & \underline{0.73} & \underline{0.88} & \underline{80.03} &  0.17 & 0.23   & 1.09   \\ 

\bottomrule
\end{tabular}

}
\end{table*}

\subsection{Quantitative Evaluation}
\noindent{\textbf{Metric Evaluation.}}
Tab.~\ref{tab:sota_comparsion} shows the quantitative results of our methods against baselines on the test set.
Our method shows substantial improvement in preserving the original behavior compared to all competitors, achieving state-of-the-art performance across all metrics.
It is notable that our method significantly outperforms competitors in ``Seg-Cons", demonstrating pixel-level alignment precision that far surpasses other approaches. 
The optimal CLIP-T and overall scores confirm that our method enhances the response to target semantics and achieves higher-quality portrait customization.

\begin{figure}[!t]
  \centering
  \includegraphics[width=1.0\linewidth]{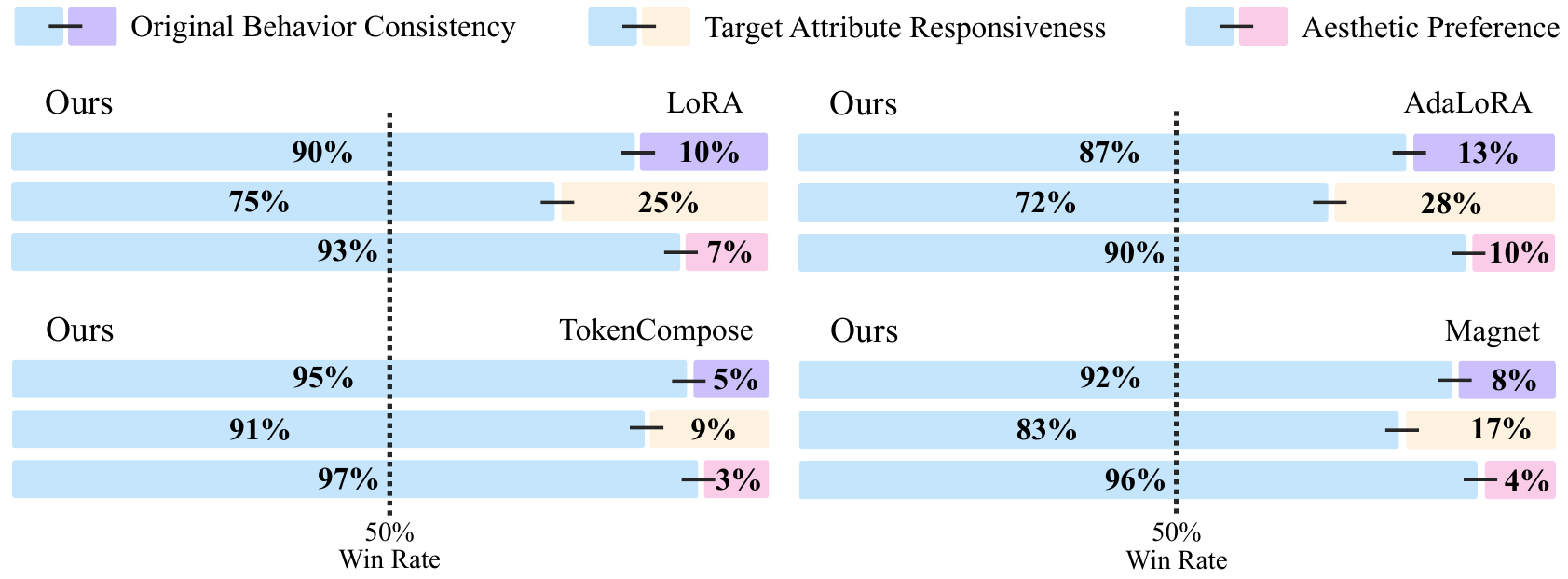}
  \vskip -1em
  \caption{\textbf{User Study Results.} The percentages indicate the proportion of users who select the corresponding method.
  }
  \vskip -1.5em
  \label{fig:user_study}
\end{figure}

\noindent{\textbf{User Study. }} We also conduct a user study to have a comprehensive assessment of our method. We design three dimensions for evaluation: Original Behavior Consistency (OBC), Target Attribute Responsiveness (TAR), and Aesthetic Preference (AP). 
We invite 32 participants from different social backgrounds, with each test session lasting about 30 minutes. Users perform pairwise comparisons between our method and competitors across three dimensions. 
The results are as shown in Fig.~\ref{fig:user_study}, our method  defeat all competitors in all dimensions, especially in OBC and TAR. This highlights our ability to preserve the original model's behavior while purely adapting to new attributes. Please refer to the \textit{Appendix} for more details.

% \vspace{-0.5em}
\subsection{Analysis of the fine-tuned model}
\begin{figure}[!t]
  \centering
  \includegraphics[width=1\linewidth]{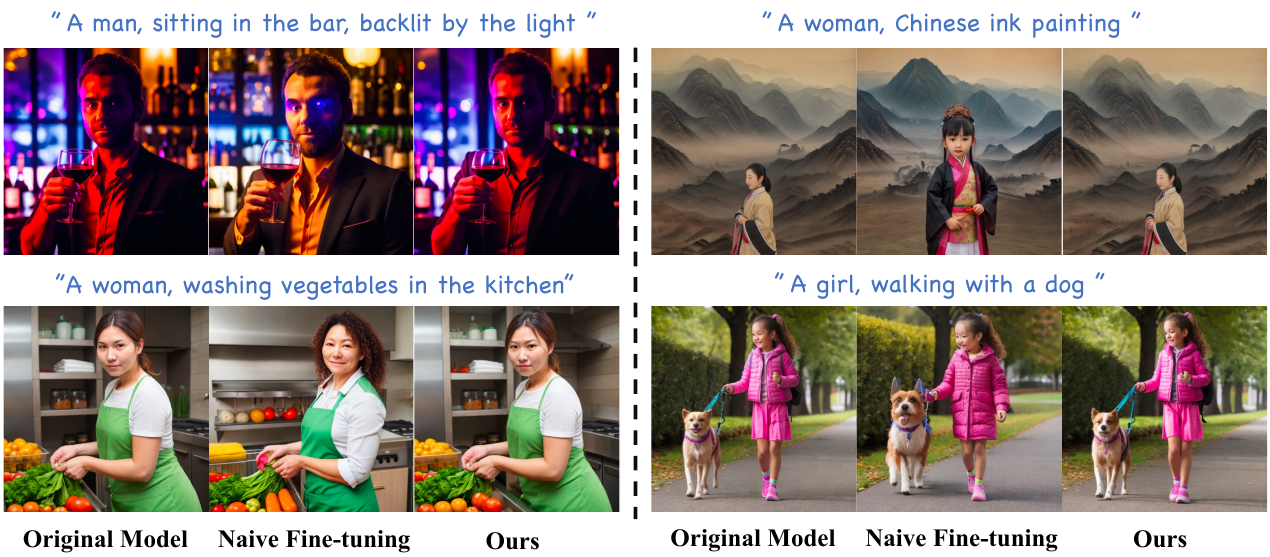}
  \vskip -1em
  \caption{\textbf{Reconstruction Results.} 
  The three portraits for each case are generated by the fine-tuned model using only the same \textit{Base text}. 
  }
  \vskip -1em
  \label{fig:reconstruction}
\end{figure}

To further verify that our method purely learns the customized attributes without compromising the original model and attains incremental learning, we solely use identical \textit{Base text} to evaluate whether our method can reconstruct the original portraits after fine-tuning.
As shown in Fig.~\ref{fig:reconstruction}, naive fine-tuning markedly disrupts original response patterns, while our method maintains near-identical performance to original model.
For example, in the top-right case, the semantics of `woman' is completely corrupted by naive fine-tuning, but we not only retain the character but also maintains high consistency in other attributes.
The outstanding reconstruction of portraits across varied scenes demonstrates our method's substantive retention of the original model's intrinsic capabilities.

\subsection{Ablation Study}
To validate the effectiveness of different components of our method, we conduct thorough ablation studies.
Qualitative results, shown in Fig.~\ref{fig:ablation}, indicate that the absence of $\mathcal{L}_{text-consistent}$ results in weaker alignment of \textit{Base text} response with the original portrait, while the lack of $\mathcal{L}_{fine-grained}$ leads to inconsistencies in detailed content, such as portrait posture. 
Without $\mathcal{L}_{enhanced}$, the expression of the target semantics significantly degrades that fails to follow the action of `holding' and with a tendency to disrupt the spatial coherence of the `toy bear', degenerating into flattened textile-like patterns.
Quantitative results in ablation part of Tab.~\ref{tab:sota_comparsion}, further validates the conclusions drawn from the visual analysis through superior performance across all metrics.
Notably, although `w/o SFCM' shows superior Preservation Metrics in Tab.~\ref{tab:sota_comparsion}, this is due to its complete disregard for target semantics and severe over-alignment with the original portrait, shown in Fig.~\ref{fig:ablation}. Such outcomes represent an absolute failure in our task, which is entirely undesirable.

\section{Conclusion}

In this paper, we propose \textbf{SPF-Portrait}, a novel fine-tuning framework designed to address the issue of \textbf{Semantic Pollution} in text-to-portrait customization.
By introducing original model as a reference path and utilizing contrastive learning, we achieve the goals of purely learning the customized semantics and enabling incremental learning.
We precisely retain the original model's behavior and ensure an effective response to target semantics by innovatively designing a Semantic-Aware Fine-Control Map to guide the alignment process and a response enhancement mechanism for target semantics.
Extensive experiments show that our method can achieve the SOTA performance.
In the future, we will continue to explore adapting our framework to more broad and complex scenes, striving to achieve semantic pollution-free fine-tuning for general text-to-image and text-to-video generation.

\bibliographystyle{ACM-Reference-Format}
\balance
\bibliography{main}

\newpage
\clearpage

\appendix

% ----------------------------------------------
% \section{\textbf{Contents of Supplementary}}
\noindent Our Supplementary Material consists of 7 sections:
\begin{itemize}
% % \vspace{+1em}
\item Section \ref{details-location} provides the training setting details of two training stages and the construction process of our training dataset.

% \vspace{+1em}
\item Section \ref{architecture-location} demonstrates that optimizing only the cross-attention layers in stage-2 can yield better performance than optimizing other architecture.
% \hyperlink{architecture-location}{\large{\textbf{C. Analysis of Fine-tuning Architecture}}}

% \vspace{+1em}
\item Section \ref{sensitivity-location}, we perform the sensitivity analysis of the hyperparameters in Eq.~\ref{eq:final_loss}.
% evidence of the hyperparamters choice in Eq.~\ref{eq:final_loss}.
% \hyperlink{sensitivity-location}{\large{\textbf{D. Sensitivity analysis of loss}}}

% \vspace{+1em}
\item Section \ref{abla_training-location} adds quantitative results of the ablation study in the training stage and demonstrates the necessity of training in two stages.
% \hyperlink{abla-location}{\large\textbf{E. Ablation Study of Training Stage}}

% \vspace{+1em}
\item Section \ref{edit-location} clarifies the fundamental distinction in task between editing methods and ours.
% \hyperlink{edit-location}{\textbf{F. Discussion about Editing Methods}}

% \vspace{+1em}
\item Section \ref{user_study-location} provides the investigation details of our user study.

% \item Visua
% \hyperlink{user-location}{\textbf{G. Details of User Study.}}
\end{itemize}

\section{Details of Our Training}
\label{details-location}

\begin{figure*}[hb]
  \centering
  \includegraphics[width=1\linewidth]{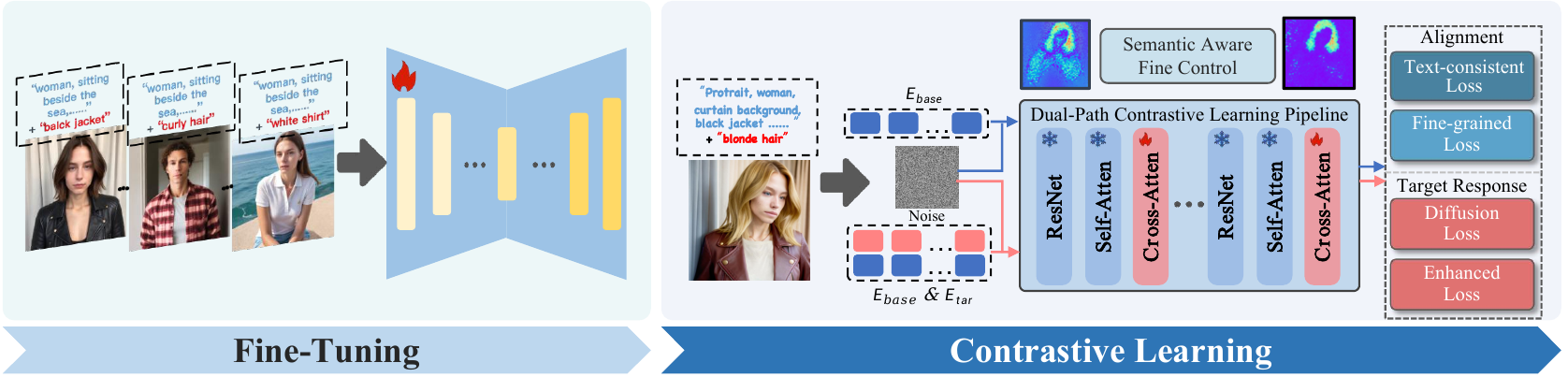}
  \caption{Our overall training pipeline.}
  \label{fig:overall}
\end{figure*}

\label{sec:details}
\subsection{Training Stage}
As shown in Fig.~\ref{fig:overall}, the training process of our approach consists of two stages: fine-tuning with all the parameters updated in the first stage and contrastive learning in the second stage. 
In the first stage, we employ conventional fine-tuning \cite{rombach2022high} to learn target attributes, which aims to enable the T2I model to rapidly adapt to the target concepts of our dataset.
For this stage, we train the model using 8 V100 GPUs with a batch size of 8, iterating for 2 epochs and a learning rate of 1e-5.
In the second stage, the training process follows the approach outlined in the main text. The goal is to enable the T2I model to grasp pure target concepts without compromising the original model’s performance, thereby preventing semantic pollution caused by the target text.
Due to the additional memory consumption of the dual-branch architecture, we set the batch size to 2, iterating for 5 epochs with a learning rate of 5e-5.
The same dataset and optimizer (AdamW with default parameters: beta1=0.9, beta2=0.999, weight decay=0.01) are used for both the first and second stages.

Due to the dual-path training framework in second stage, which requires an additional frozen original model compared to standard fine-tuning, our approach incurs extra memory costs and increased computation time. 
We provide the corresponding resource consumption details in Tab.~\ref{tab:cost}. The frozen model (which doesn't participate parameter updates) adds only 5GB of GPU memory overhead under typical FP32 precision settings.

\begin{table}[ht]
\centering
\caption{\textbf{Computation Time and Memory Usage of Training under Different Data Type.} The data in bold represents our implementation configuration.
}
\label{tab:cost}
\resizebox{\linewidth}{!}
{
\begin{tabular}{lccc}
\toprule
\multicolumn{1}{c}{\multirow{2}{*}{\textbf{Method}}} & \multicolumn{3}{c}{\textbf{Data Type}} \\
 \cmidrule(lr){2-4}
 & FP16 & FP32  &BP16 \\
\midrule
Stage-1 (w/o reference pat) & 1.92s/iter (17GB) & \textbf{2.28s/iter (23GB)} & OOM \\
Stage-2 (w/ reference path) & 2.1s/iter (21GB) & \textbf{3.26s/iter (28GB)}   & OOM\\
\bottomrule
\end{tabular}
}
\end{table}

\subsection{Training Dataset}
\begin{figure*}[!ht]
  \centering
  \includegraphics[width=1\linewidth]{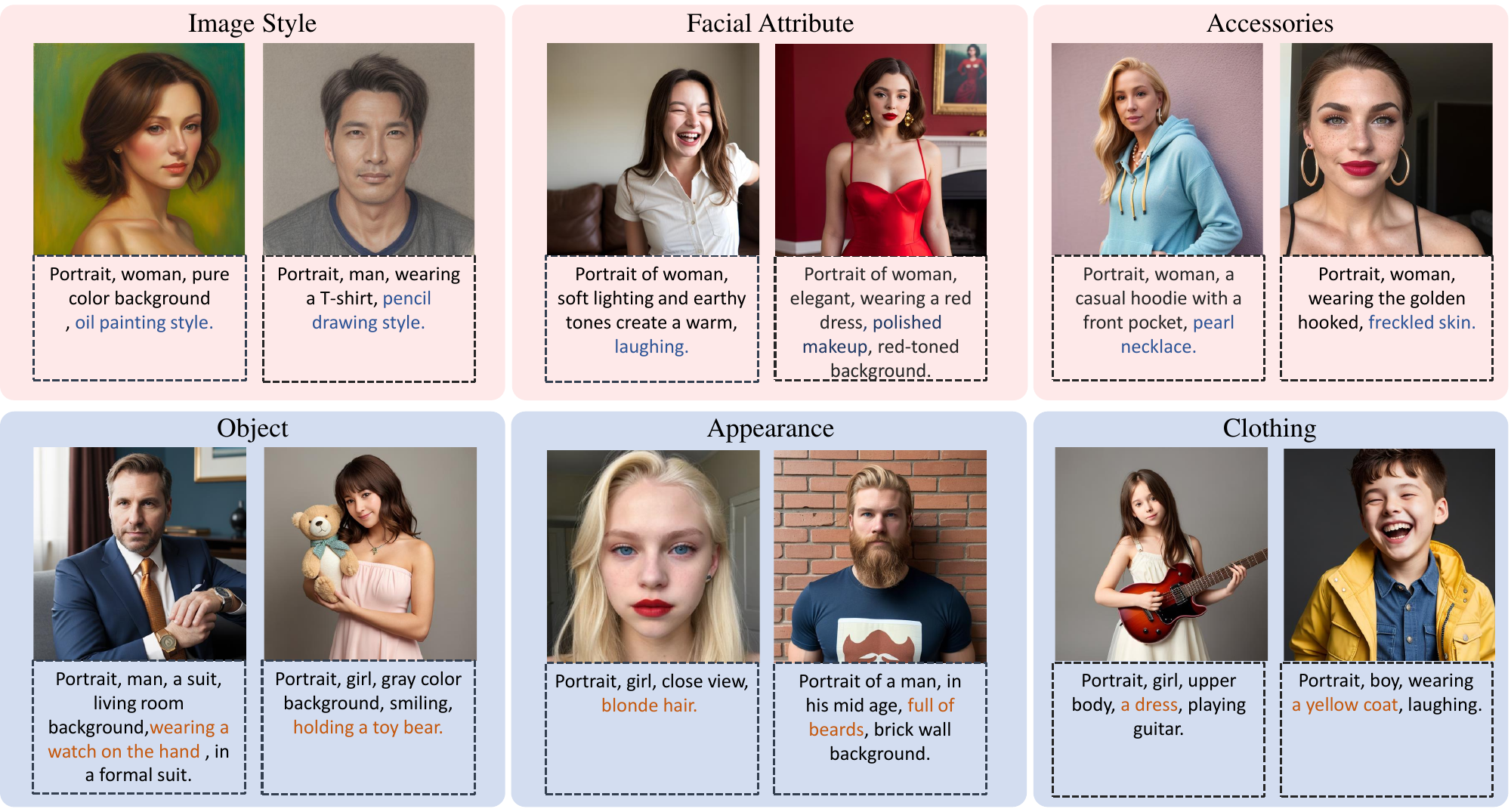}
  \caption{\textbf{Example of our training datasets.}
  }
  \label{fig:dataset}
\end{figure*}

Our work focuses on preventing semantic pollution in fine-tuning portrait T2I models while enabling the model to learn the concepts from the target attributes.
To achieve this, we constructed a dataset containing various image-text pairs related to portrait concepts for training the T2I diffusion model. 
Considering the quality and diversity of the dataset, we utilized widely adopted community checkpoints for portrait generation as the checkpoints for the Stable Diffusion (SD) model, including RealVisXL\_V1.0 and HumanModel, to generate portrait images encompassing a wide range of attributes.
The attribute statistics and corresponding samples are shown in Tab.~\ref{tab:dataset} and Fig.~\ref{fig:dataset}, respectively.

To improve dataset quality, we focus on two aspects: 1) enhancing image-text alignment using FLIP \cite{li2024flip}, a CLIP checkpoint specifically for portraits, to retain the top 30\% of matching pairs, and 2) improving visual fidelity by filtering images with a Human Aesthetic Preference Score (HPS) and Image-Reward (IR).

\begin{table}[h]
\centering
\caption{\textbf{Details of Our Training Dataset.} The specific categories of character attributes covered by our training dataset.}%
\resizebox{0.5\linewidth}{!}{
\begin{tabular}{@{}c|cc@{}}
\toprule
Category        & number   \\ \midrule
Facial Attributes & 52021         \\
Clothing       & 67871         \\
Image Style       & 36786         \\
Appearance       & 27508        \\
Accessories       & 45200         \\\bottomrule
\end{tabular}
}
\label{tab:dataset}
\end{table}

\begin{figure*}[ht]
  \centering
  \includegraphics[width=1\linewidth]{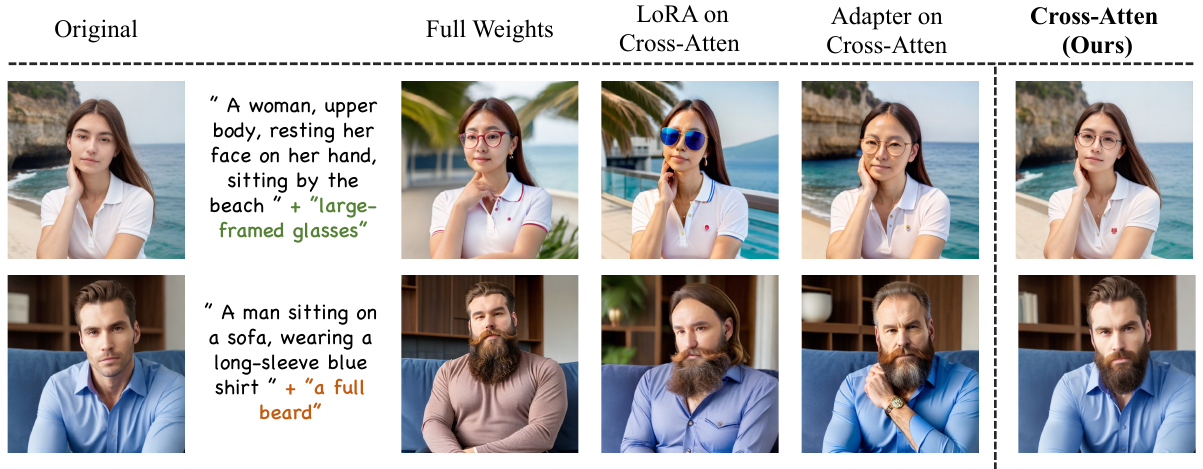}
  \caption{\textbf{Comparison of results across different updated network architectures in our constraive pipeline.}
  "Full Weights" indicates that all network parameters are updated, "LoRA on Cross-Atten" refers to the integration of LoRA into the Cross-Attention modules, and "Adapter on Cross-Atten" denotes the addition of parallel cross-attention layers, akin to IP-adapter \cite{ye2023ip}.}
  \label{fig:architecture}
\end{figure*}

% \clearpage

\section{Analysis of Fine-tuning Architecture.}
\label{architecture-location}

\begin{table*}[ht]
\centering
\caption{
\textbf{Quantitative Comparisons} with other architecture. 
}
\label{tab:architecture}
\resizebox{1\textwidth}{!}
{
\begin{tabular}{lcccccccc}
\toprule
\multicolumn{1}{c}{\multirow{2}{*}{\textbf{Method}}} & \multicolumn{5}{c}{\textbf{Preservation}} & \multicolumn{2}{c}{\textbf{Overall}} & {\textbf{Responsiveness}}\\
\cmidrule(r){2-6} \cmidrule(r){7-8} \cmidrule(r){9-9}
& FID ($\downarrow$) & LPIPS ($\downarrow$) & ID ($\uparrow$) & CLIP-I ($\uparrow$) & Seg-Cons ($\uparrow$) & HPSv2 ($\uparrow$) & MPS($\uparrow$)   & CLIP-T ( $\uparrow$)  \\ 
\midrule
\multicolumn{1}{l|}{Full Weights}           & 7.82  & 0.40   &0.309  &0.81   &48.39  &0.22   &0.87  &0.26\\
\multicolumn{1}{l|}{LoRA on Cross-Atten}    & 7.10  & 0.39   &0.487  &0.61   &68.37  &0.24   &1.21  &0.26\\
\multicolumn{1}{l|}{Adapter on Cross-Atten} & 5.93  & 0.37   &0.520  &0.80   &61.70  &0.25   &1.31  &0.27\\
\midrule
\rowcolor{tb-blue!40}\multicolumn{1}{l|}{\textbf{Ours}} & \textbf{4.50} & \textbf{0.35} & \textbf{0.55} & \textbf{0.83} & \textbf{75.74} & \textbf{0.28} & \textbf{1.49} & \textbf{0.30} \\

\bottomrule
\end{tabular}
}
\end{table*}

During the contrastive learning of the second stage, our approach exclusively trains the parameters in the cross-attention modules. 
We compare results across various network architectures, including "full-weight", "LoRA on cross-attention", and "additional adapters". As illustrated in Fig.~\ref{fig:architecture}, all architectures under our contrastive learning achieve some level of alignment. Notably, "LoRA on cross-attention", "Adapter on Cross-Atten" and "Cross-Atten(ours)" outperform the "full weights" in alignment, this is because the diffusion model relies on the cross-attention mechanism for text-conditioned control, and optimizing the most critical parameters enables a better understanding of independent target attributes. 
However, "LoRA on Cross-Atten", due to its limited learnable parameters, falls short in understanding the original behavior compared to our method. Ours achieves a superior balance between alignment and attribute learning.
"Adapter on Cross-Atten" achieves the suboptimal performance, as it independently adjusts all the parameters of cross-attention module. However, the isolated attention structure limits the interaction between target text features and base text features, rendering in partial misalignment.
The results in Tab.~\ref{tab:architecture} further validate our conclusions.

% \clearpage

\section{Sensitivity Analysis of Loss}
\label{sensitivity-location}
\begin{figure*}[hb]
  \centering
  \includegraphics[width=1.0\linewidth]{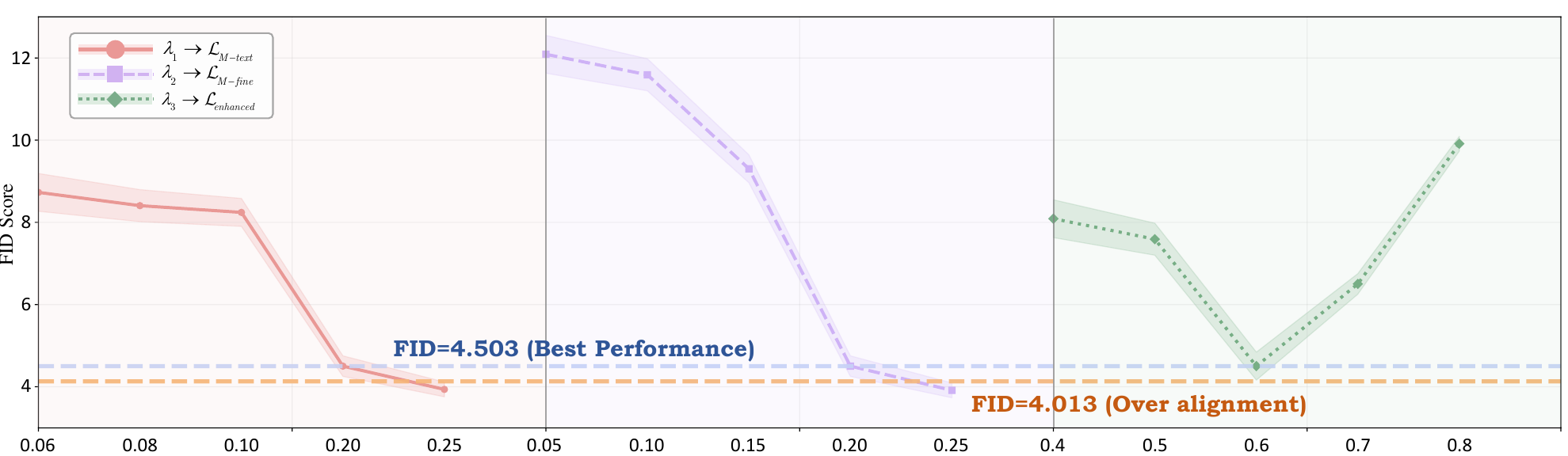}
  \caption{\textbf{Sensitivity analysis of three loss components ($\lambda_{1} \rightarrow \mathcal{L}_{M-text}$, $\lambda_{2} \rightarrow \mathcal{L}_{M-fine}$, $\lambda_{3} \rightarrow \mathcal{L}_{M-enhanced}$) with respect to FID scores.}. FID varies with different parameter values for each loss component. FID=4.503 (optimal performance) and FID=4.013 (over-alignment).
  }
  \label{fig:sensitivity}
\end{figure*}

To determine the optimal settings for the three loss hyperparameters, we conducted a comprehensive sensitivity analysis. As shown in Fig. \ref{fig:sensitivity} The three segments of the plot correspond to the hyperparameters in Eq. 11 ($\lambda_{1} \rightarrow \mathcal{L}_{M-text}$, $\lambda_{2} \rightarrow \mathcal{L}_{M-fine}$, $\lambda_{3} \rightarrow \mathcal{L}_{M-enhanced}$) , demonstrating how FID scores vary with their values. Our analysis reveals that the optimal configuration occurs at $\lambda_{1} =0.2, \lambda_{2} =0.1, \lambda_{3}=0.6$, achieving the best FID score of 4.503 reported in our main results. It's noticed that the orange dashed line indicates the FID (4.013) of "Ours(w/o SFCM)" from Tab.~\ref{tab:sota_comparsion}, which exhibits over-alignment as visualized in Fig.~\ref{fig:ablation}.

\section{Ablation Study of Training Stage}
\label{abla_training-location}
The main contribution of our method is the addition of an extra training stage on top of naive fine-tuning. To demonstrate the effectiveness of the two-stage training strategy, we conduct an ablation study on the training stages. As shown in Tab. \ref{tab:training_stage}, if only the second-stage contrastive learning is used, the model struggles to learn clean target attributes, resulting in significantly poor performance on "CLIP-T." On the other hand, with only stage 1, the model is entirely affected by semantic pollution, failing to align with the original model behavior, thus performing worse on preservation metrics.

\begin{table*}[ht]
\centering
\caption{\textbf{Ablation Study of the training Stage.}}
\label{tab:training_stage}
\begin{tabular}{lcccccccc}
\toprule
\multicolumn{1}{c}{\multirow{2}{*}{\textbf{Method}}} & \multicolumn{5}{c}{\textbf{Preservation}} & {\textbf{Responsiveness}} & \multicolumn{2}{c}{\textbf{Overall}} \\
\cmidrule(r){2-6} \cmidrule(r){7-7} \cmidrule(r){8-9} 
& FID ($\downarrow$) & LPIPS ($\downarrow$) & ID ($\uparrow$) & CLIP-I ($\uparrow$) & Seg-Cons ($\uparrow$) &  CLIP-T ( $\uparrow$) & HPSv2 ($\uparrow$)   & MPS($\uparrow$)  \\ 
\midrule
\multicolumn{1}{l|}{Only Stage-1 (Naive Fine-tuning)} & 20.41   & 0.57 & 0.21  & 0.63  & 57.77 & 0.24 & 0.21 & 0.67 \\
\multicolumn{1}{l|}{Only Stage-2} & 7.18 & 0.38 & 0.13 & 0.71 & 63.72 & 0.19 & 0.24 & 1.12 \\
\midrule
\rowcolor{tb-blue!40}\multicolumn{1}{l|}{\textbf{Stage-1\&2 (Ours)}}  & \textbf{4.50} & \textbf{0.35} & \textbf{0.55} & \textbf{0.83} & \textbf{75.74} & \textbf{0.30} & \textbf{0.28} & \textbf{1.49} \\
\bottomrule
\end{tabular}
\end{table*}

\section{Discussion about Editing Methods}
\label{edit-location}
As mentioned in the related work Sec.~\ref{sec:relat        ed_work}, incorporating text-driven editing methods \cite{deutch2024turboedit, wang2024belm, kim2022diffusionclip, ju2024brushnet} into the T2I model pipeline can produce similar results to ours.
Here, we elaborate on the distinctions between our work and editing models and demonstrate that the improvement on inversion-based editing models when replacing their T2I model with ours.

The core distinction of our work lies in preventing additional textual concepts from disrupting T2I models, which fundamentally differs from I2I editing models that primarily focus on image manipulation through precise local modifications.
Although the visual results of our method are presented in a pairwise comparison which may resemble those of editing work, the purpose is to demonstrate that our incremental learning approach preserves the integrity of the original model.

For an ideal AI-driven text-to-portrait creation, users aim for text to function like a brush in traditional painting, enabling targeted modifications to specific regions while preserving others unchanged.
With existing technology, users can only achieve this by combining text-driven editing models, requiring:
1) Initial creation using a T2I model, 2) Refinement with an I2I editing model.
However, in our framework, the T2I model can directly modify images via controlled text input during continuous generation, eliminating the need for additional I2I editing models.
It can maintain consistency across continuous generations by preserving identical content for shared text elements.
This makes the creative process more controllable, convenient, and aligned with intuition.

\section{Details of User study}
\label{user_study-location}
\begin{figure*}[!h]
  \centering
  \includegraphics[width=1.0\linewidth]{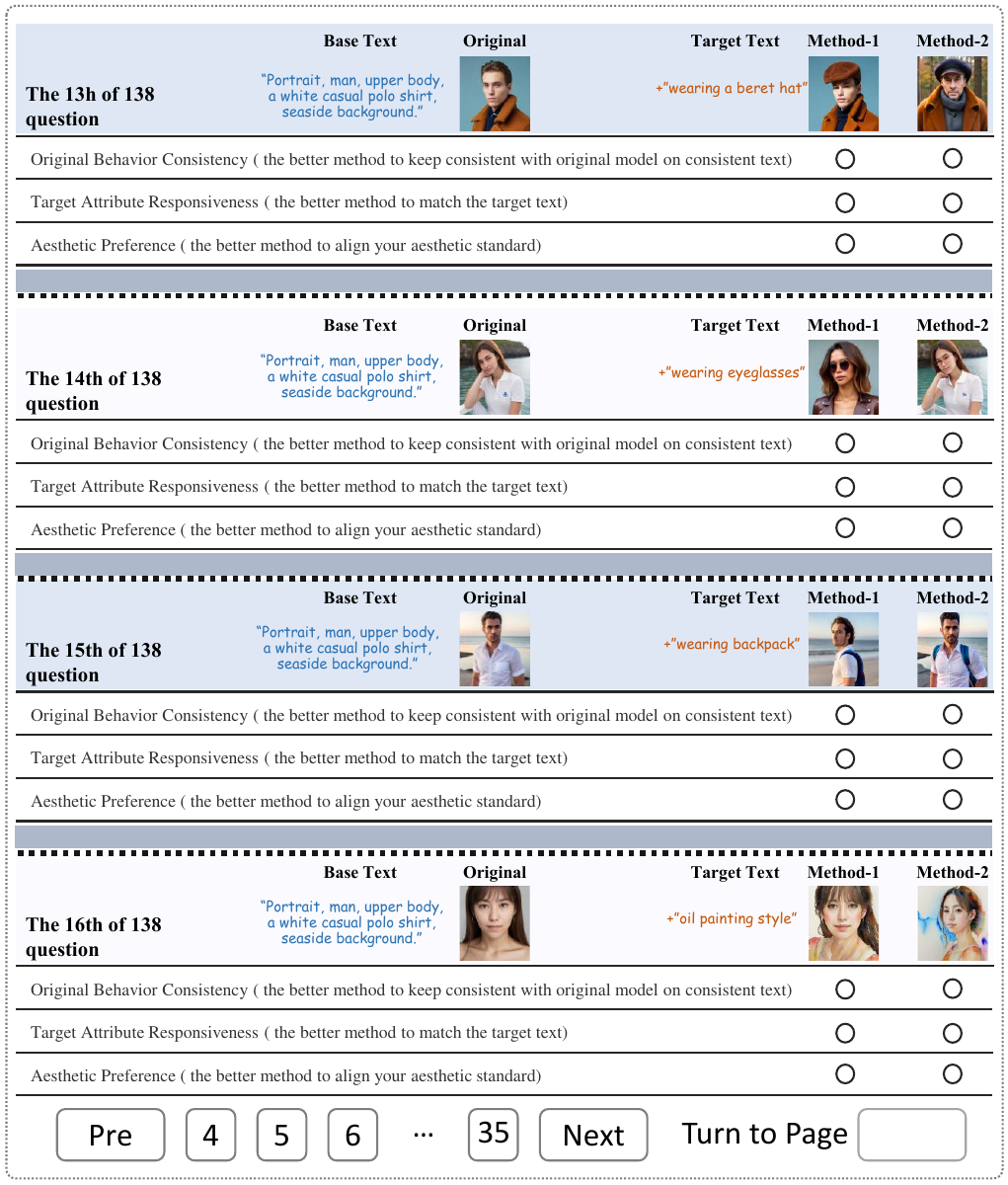}
  \caption{\textbf{The investigation page in user study.}
  }
  \label{fig:user}
\end{figure*}

We provide more details on our user study implementation.
Besides qualitative and quantitative comparisons, we also conduct a user study to determine whether our method is preferred by humans and to underst and how people perceive emotions. We invite 32 participants from different social backgrounds and each test session lasts about 30 minutes. 
During the investigation, as illustrated in Fig. \ref{fig:user}, we conducted a pairwise comparison between our method and competitors across three key dimensions: Original Behavior Consistency, text alignment, and human preference.
For "Original Behavior Consistency", users were asked to select which of the two images better preserved consistency with the original model’s outputs. 
For "Target Attribute Response", users evaluated which image more accurately reflected the target text description. 
For "Aesthetic Preference", users judged which image aligned better with their aesthetic preferences, considering factors such as visual quality and the absence of artifacts or distortions.
This comprehensive evaluation framework ensures a thorough and objective assessment of our method’s performance relative to existing approaches.
The generation results are evaluated on three dimensions: image fidelity, text alignment, and human preference.

\begin{figure*}[tb]
  \centering
  \includegraphics[width=0.92\linewidth]{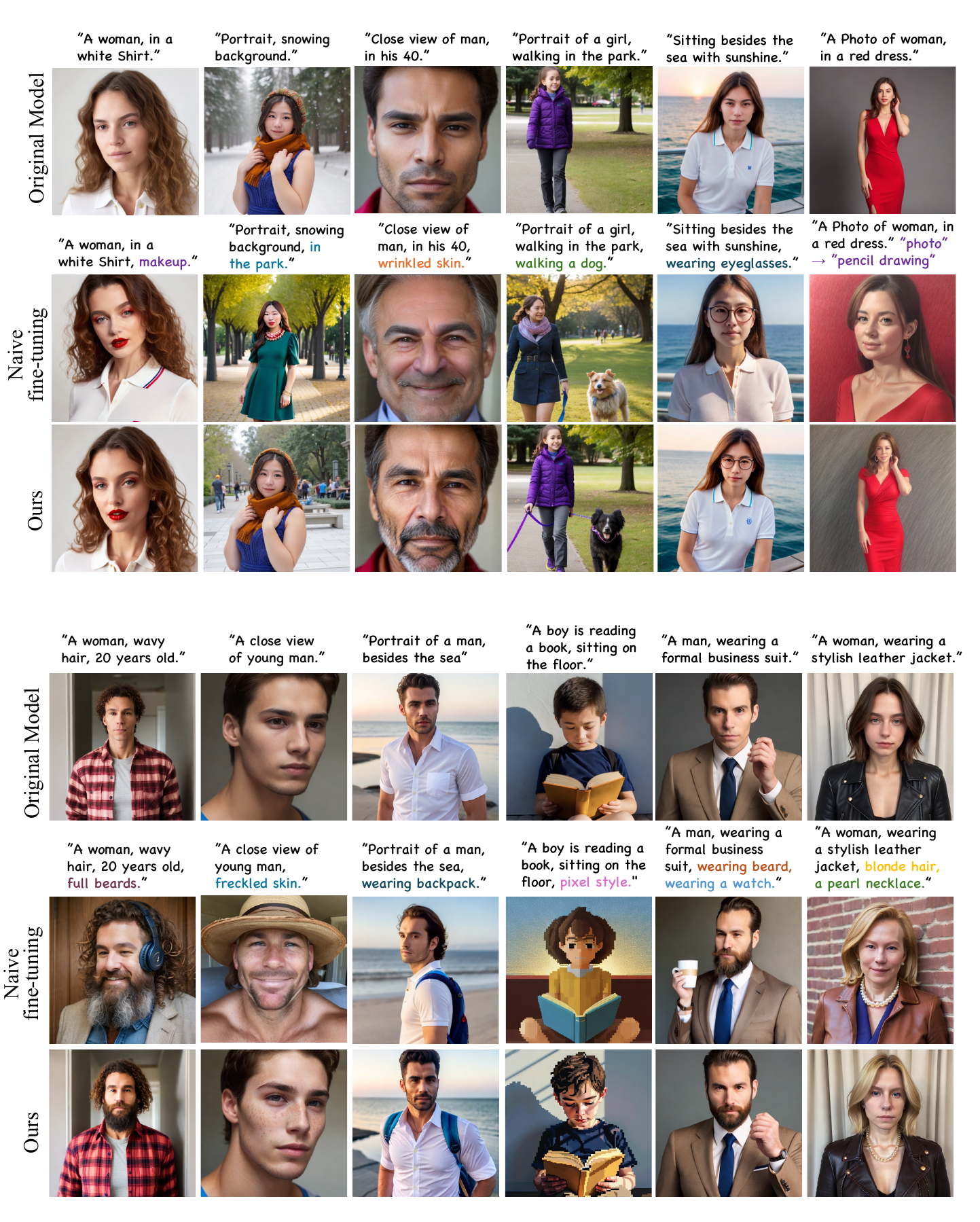}
  \caption{\textbf{More Results of SPF-Portrait on Text-to-Portrait Customization.} 
  }
  \label{fig:appendix_sota}
\end{figure*}

\begin{figure*}[tb]
  \centering
      \includegraphics[width=0.95\linewidth]{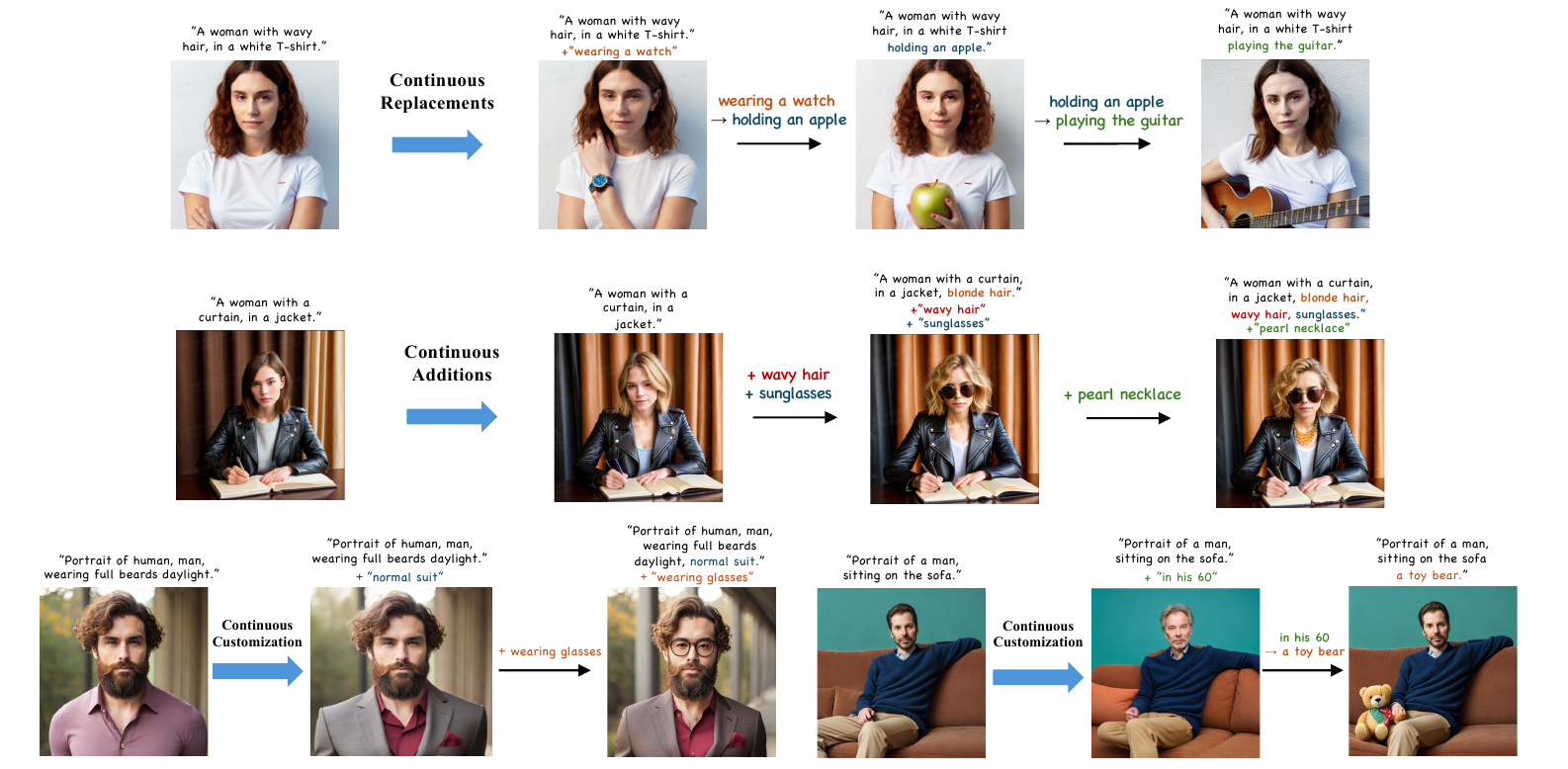}
  % \vskip -1em
  \caption{\textbf{Results of continuous replacements and additions of target text
in text-to-portrait customization. } Our method demonstrates stable and excellent performance in continuous customization tasks, indicating its potential to play a role in the application scenarios of continuous AI creation.
  }
  \label{fig:appendix_con}
\end{figure*}

\begin{figure*}[tb]
  \centering
  \includegraphics[width=0.95\linewidth]{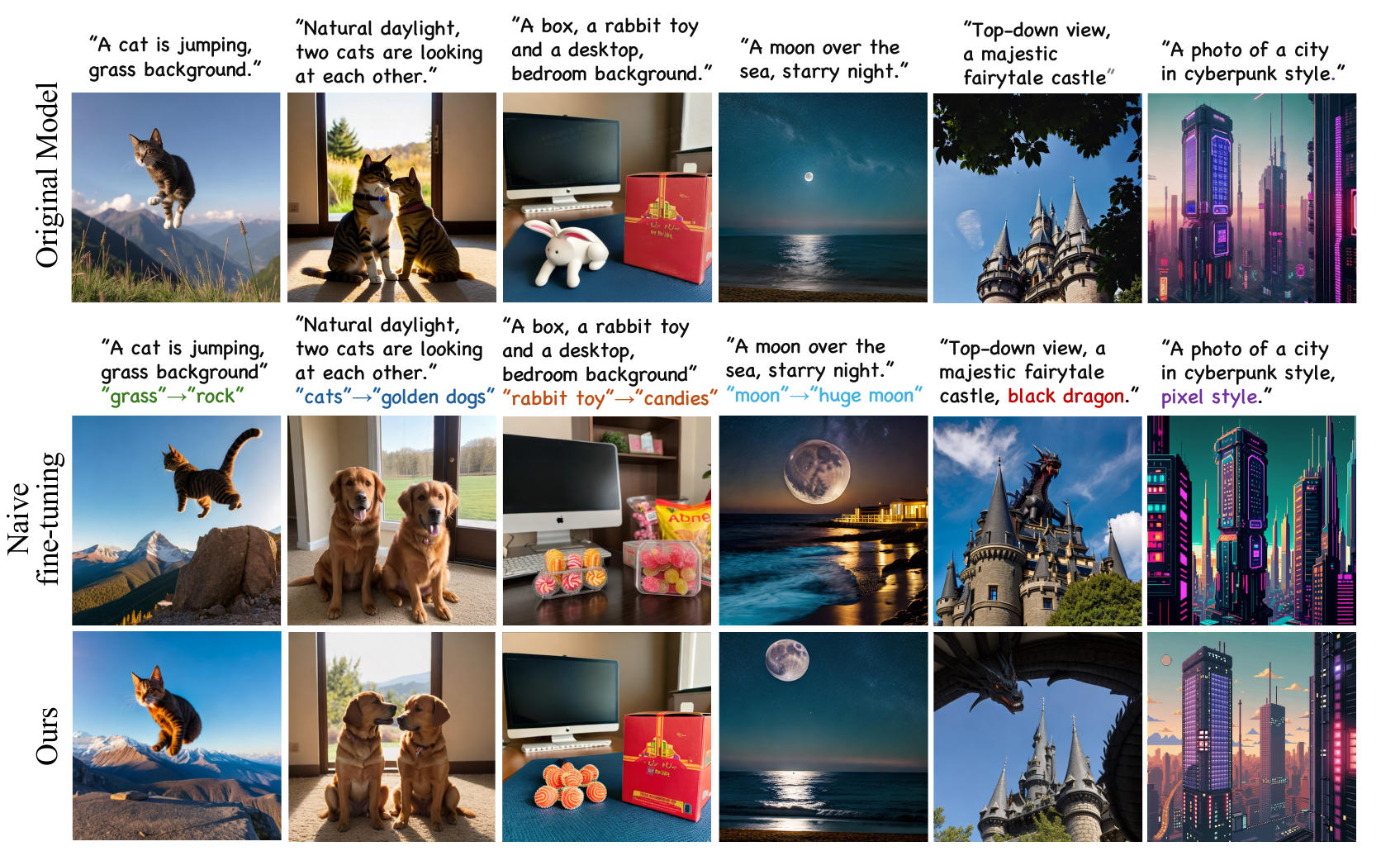}
  % \vskip -1em
  \caption{\textbf{Results of extending our method to the general Text-to-Image domain. }These excellent experimental results demonstrate the feasibility of extending our method to the general T2I domain. Our method has the potential to address the issue of semantic pollution in fine-tuning and to achieve incremental learning within the general T2I domain.
  }
  \label{fig:appendix_gs}
\end{figure*}

\clearpage

\end{document}